\pdfoutput=1

\documentclass[11pt]{article}

\usepackage[final]{acl}
\usepackage{fc_custom}

\usepackage{times}
\usepackage{latexsym}
\usepackage{multirow}

\usepackage[T1]{fontenc}

\usepackage[utf8]{inputenc}

\usepackage{microtype}
\usepackage{booktabs}
\usepackage{graphicx}
\usepackage{diagbox}
\usepackage{longtable}
\usepackage{tabularx}
\usepackage{tabularray}
\usepackage{adjustbox}
\usepackage{amsmath}
\usepackage{amssymb}
\usepackage{algorithm}
\usepackage{algorithmicx}
\usepackage{algpseudocode}
\usepackage{rotating}
\newtheorem{theorem}{Theorem}

\title{\ti{GeoHard}: Towards Measuring Class-wise Hardness through \\ Modelling Class Semantics}

\author{
\textbf{Fengyu Cai}\textsuperscript{1} \,
\textbf{Xinran Zhao}\textsuperscript{2} \,
\textbf{Hongming Zhang}\textsuperscript{3} \,
\textbf{Iryna Gurevych}\textsuperscript{1} \,
\textbf{Heinz Koeppl}\textsuperscript{1}
\vspace{5pt} \\ 
\textsuperscript{1}Technical University of Darmstadt \,
\textsuperscript{2}Carnegie Mellon University \,
\textsuperscript{3}Tencent AI Lab \\
\texttt{\{fengyu.cai, heinz.koeppl\}@tu-darmstadt.de}
}

\begin{document}
\maketitle

\begin{abstract}
Recent advances in measuring hardness-wise properties of data guide language models in sample selection within low-resource scenarios.
However, class-specific properties are overlooked for task setup and learning.
How will these properties influence model learning and is it generalizable across datasets?
To answer this question, this work formally initiates the concept of \ti{class-wise hardness}. 
Experiments across eight natural language understanding (NLU) datasets demonstrate a consistent hardness distribution across learning paradigms, models, and human judgment.
Subsequent experiments unveil a notable challenge in measuring such class-wise hardness with instance-level metrics in previous works.
To address this, we propose \ti{GeoHard} for class-wise hardness measurement by modeling class geometry in the semantic embedding space.
\ti{GeoHard} surpasses instance-level metrics by over 59 percent on \ti{Pearson}'s correlation on measuring class-wise hardness. 
Our analysis theoretically and empirically underscores the generality of \ti{GeoHard} as a fresh perspective on data diagnosis.
Additionally, we showcase how understanding class-wise hardness can practically aid in improving task learning.
The code for \ti{GeoHard} is available \footnote[1]{\url{https://github.com/TRUMANCFY/geohard}}.

\end{abstract}

\section{Introduction}
Data acts as a crucial intermediary proxy for AI systems to understand and tackle real-world tasks \cite{DBLP:conf/cvpr/TorralbaE11,DBLP:journals/corr/abs-1811-12569}.
Therefore, evaluating the hardness of individual instances, or instance-level hardness \cite{DBLP:conf/emnlp/KongJZLZZ20,DBLP:journals/tacl/HahnJF21,DBLP:conf/icml/EthayarajhCS22,DBLP:conf/emnlp/ZhaoMM22}, relative to the dataset is key for learning and analyzing NLP tasks.
This evaluation is increasingly important with the rise of large language models (LLMs; \citealt{DBLP:journals/corr/abs-2307-09288,chung2024scaling}). Measuring hardness aids in selecting examples for in-context learning (ICL; \citealt{ye2024investigating}) or training samples for fine-tuning models \cite{DBLP:journals/corr/abs-2305-11206, xie2023data}.

\begin{figure}[t]
		\centering
		\includegraphics[width=0.8\columnwidth]{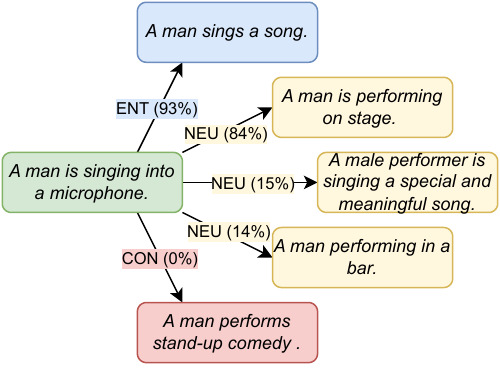}
		\caption{The examples of premise-hypothesis pairs in uncertain NLI ($u$-NLI; \citealt{DBLP:conf/acl/ChenJPSD20}). In $u$-NLI, the probability of these pairs (in the parentheses) is annotated by crowdworkers.
            The example showcases \ts{NEU}'s \ti{Middlemost} and \ti{Diverse} semantics, i.e., positioning in the middle between \ts{ENT} and \ts{CON} and widely ranging from low (14\%) to high probability (84\%).}
		\label{fig:unli}
\end{figure}

However, another critical yet underexplored component of the dataset is the classes themselves, whose properties, such as ambiguity in their definitions, can also contribute to difficulties. While considerable efforts have been made to address class imbalance in specific datasets \cite{DBLP:conf/emnlp/Subramanian0BCF21, DBLP:conf/eacl/HenningBFF23}, there remains a lack of comprehensive analysis on class-wise properties that are consistent across different tasks.
Conventionally, classes are treated equally, e.g., the demonstrations in In-Context Learning (ICL) typically being evenly sampled among classes \cite{DBLP:conf/emnlp/MinLHALHZ22}.
This raises an important question: \textit{How do class-specific properties influence model performance?}

We formally initiate the concept of \ti{class-wise hardness} as the relative difficulty of a class, in analogy to instance-level hardness \cite{DBLP:conf/icml/EthayarajhCS22}.
To make this notion quantifiable, we present the concept of the \ti{empirical} class-wise hardness which assesses the class-specific performance given an LM and learning paradigm.
Subsequently, the intrinsic class-wise hardness can be approximated by pooling the empirical performances across models and learning paradigms.
Our analysis across eight Natural Language Inference (NLI) or Sentiment Classification (SC) tasks reveals the consistent challenge of \ti{Neutral} across a spectrum of tasks, learning paradigms, and models together with human annotation disagreement \cite{DBLP:conf/emnlp/NieZB20}.
These findings verify the concept and establish the estimation of inherent class hardness.

Then, we study how to measure these class-specific properties leading to consistent class-wise hardness.
We first show that naively aggregating Sensitivity Analysis (SA, \citealt{DBLP:journals/tacl/HahnJF21}) and two similarity-based methods \cite{DBLP:conf/emnlp/ZhaoMM22,DBLP:journals/corr/abs-2307-10442} fails in measuring class-wise hardness across datasets.
This stimulates us to propose a specific metric for class-wise hardness measurement beyond the instance-level measurement.
We propose an effective, lightweight, and training-free metric, \ti{GeoHard}, which analyzes data distribution from the geometrical space of semantic embeddings.
\ti{GeoHard} utilizes both \ti{inter-} and \ti{intra-}class properties, e.g., \ti{Neutral}'s \ti{MiddleMost} and \ti{Diverse} semantics shown in Figure \ref{fig:unli}, respectively.
Our experiments show that \ti{GeoHard} demonstrates its exceptional capacity in measuring class-wise hardness, outperforming the instance-level aggregation by over 59 percent on \ti{Pearson}'s correlation between measurement and reference.
Our theoretical and experimental analysis validates its generalization to other tasks without further adaptation.

As for the practical perspective, we show how to use \ti{GeoHard} to improve \ti{task learning} with class reorganization \cite{nighojkar-etal-2023-strong}.
Class reorganization targets a balanced class performance, e.g., by splitting one hard class into two sub-classes \cite{DBLP:conf/acl/PottsWGK20}.
\ti{GeoHard} is shown to be able to well interpret the heuristic-based reorganization proposed in the previous work \cite{DBLP:conf/acl/PottsWGK20}.
We demonstrate that class-aware demonstration selection guided by \ti{GeoHard} also benefits ICL.

Our contribution is three-fold:
\begin{enumerate}[noitemsep,labelwidth=*,leftmargin=1.8em,align=left]
    \item We initiate the concept of class-wise hardness (Section \ref{sec:hardness}) and show that the direct aggregation of the current instance-level hardness metrics fails to correlate with class-wise hardness on 8 NLI/SC datasets (Section \ref{sec:exp});
    \item We instead target class semantics and put forward a geometry-based method, \ti{GeoHard}, which outperforms the baselines by 59\% (Section \ref{sec:geohard}). We theoretically and empirically show \ti{GeoHard}'s promising generalization to other tasks (Section \ref{sec:generalization});
    \item We demonstrate the potential application of class-wise hardness measured by \ti{GeoHard} to interpret class reorganization and improve task learning (Section \ref{sec:application}).
\end{enumerate}

\section{Formulation of Class-wise Hardness} \label{sec:hardness}
Here, we define class-wise hardness as the difficulty of the class across all the classes, akin to instance-level hardness \cite{DBLP:conf/icml/EthayarajhCS22}.
Formally, given the classes $\mathcal{C} = \{c_1, ..., c_K \}$ for a classification task where $c_k$ is a class, $c_k$'s class-wise hardness can be denoted as $\mathrm{H}(c_k \mid \mathcal{C})$.
We denote $\mathrm{\tf{H}}(\mathcal{C}) = [ \mathrm{H}(c_1 \mid \mathcal{C})), ..., \mathrm{H}(c_K \mid \mathcal{C}) ]$.

As $\mathrm{H}$ is intractable, we can empirically obtain class-wise hardness by assessing the performance of $c_k$ given the LM $m \in \mathcal{M} $, e.g., Flan-T5-Large \cite{DBLP:journals/jmlr/RaffelSRLNMZLL20} or LLaMA-2-13B \cite{DBLP:journals/corr/abs-2307-09288}, and learning paradigms $l \in \mathcal{L}$, e.g., fine-tuning or ICL.
We denote this empirical class-wise hardness conditioned on LMs and learning paradigms as $\tilde{\mathrm{H}}(c_k \mid \mathcal{C}, m, l)$.
Therefore, class-wise hardness $\mathrm{H}$ can be approximated by marginalizing $\tilde{\mathrm{H}}$ on the pairs of models and learning paradigms 
$\mathcal{P} = \{ (m, l) \mid m \in \mathcal{M}, l \in \mathcal{L} \}$:

\begin{align}
    \mathrm{H}(c_k \mid \mathcal{C}) &= \mathbb{E}_{(m, l) \in \mathcal{P}}[\tilde{\mathrm{H}}(c_k \mid \mathcal{C}, m, l)] \\
    &\approx \frac{ \sum_{\substack{(m, l) \in \mathcal{P}}} \tilde{\mathrm{H}}(c_k \mid \mathcal{C}, m, l)}{| \mathcal{P} |}\label{eq:margin}
\end{align}

In the rest of this section, we calculate empirical class-wise hardness on eight NLI/SC datasets.
We observe the consistency of $\tilde{\mathrm{\tf{H}}}$ among LMs, learning paradigms, and human annotation, which stimulates us to simplify the approximation of $\mathrm{H}$.

\subsection{Datasets}
We initiate class-wise hardness with 8 NLU datasets, comprising 3 NLI datasets and 5 SC datasets, as shown in Table \ref{tab:dataset} and Table \ref{tab:appendix:dataset} in Appendix \ref{appendix:data_examples}.
We chose these datasets based on their popularity and their similar format for comparison.
We normalize the label format of the SC datasets to \ti{Positive}, \ti{Neutral}, and \ti{Negative}, as described in Appendix \ref{appendix:data_norm}.
Lastly, we balance the number of instances within each class\footnote[2]{We eliminate the potential influence of class imbalance by randomly sampling the same number of instances belonging to each class in training, validation, and test sets, respectively.}.
Class imbalance is shown to negatively affect the performance of minority classes \cite{DBLP:conf/eacl/HenningBFF23}.

\subsection{Calculation of Empirical Hardness $\tilde{\mathrm{H}}$} \label{sec:hardness_LMs}
To achieve a precise and complete approximation on $\mathrm{H}$, we encompass various pairs of LMs and learning paradigms for the calculation of $\tilde{\mathrm{H}}(c_k \mid \mathcal{C}, m, l)$, as outlined in Equation \ref{eq:margin}.

\paragraph{Inter-annotator disagreement} can reflect the difficulty of the instance, namely that the higher human disagreement implies more hardness on data \cite{DBLP:conf/emnlp/NieZB20, basile-etal-2021-need}.
Hence, we calculate class-wise human disagreement as the average entropy of the annotation distribution of the instances labeled on MNLI and SNLI \footnote[3]{Only the inter-annotator agreement of MNLI and SNLI is evaluated due to data accessibility.}.
Referring to Table \ref{tab:human_disagreement} in the Appendix \ref{appendix:hardness:human}, \ti{Neutral}'s class-wise human disagreement is the highest, indicating its exceptional hardness w.r.t. human.

\begin{table}[t]
\centering
\setlength{\tabcolsep}{8pt}
{\scriptsize
\begin{tabular}{lccc}
\toprule
\multicolumn{1}{c}{\%}  & \tf{Roberta-Large} & \tf{OPT-350M} & \tf{Flan-T5-Large} \\ \midrule
{Amazon} &87.6\mmid\tf{71.0}\mmid80.6 & 87.0\mmid\tf{68.7}\mmid79.3 & 88.6\mmid\tf{71.6}\mmid81.3 \\
{APP} &74.2\mmid\tf{60.1}\mmid73.4 & 73.6\mmid\tf{56.1}\mmid72.6 & 74.3\mmid\tf{59.0}\mmid73.9 \\
{MNLI} &91.0\mmid\tf{87.2}\mmid92.9 & 86.1\mmid\tf{80.5}\mmid85.8 & 91.3\mmid\tf{87.5}\mmid92.9 \\
{SICK-E} &92.9\mmid\tf{86.8}\mmid92.4 & 85.8\mmid\tf{79.1}\mmid89.1 & 92.9\mmid\tf{85.7}\mmid92.4 \\
{SNLI} &92.6\mmid\tf{89.2}\mmid95.3 & 91.0\mmid\tf{86.5}\mmid92.3 & 92.8\mmid\tf{89.7}\mmid95.5 \\
{SST-5} &83.1\mmid\tf{53.1}\mmid75.8 & 82.3\mmid\tf{55.6}\mmid71.5 & 83.4\mmid\tf{51.7}\mmid76.1 \\
{TFNS} &93.0\mmid\tf{86.1}\mmid92.2 & 88.0\mmid\tf{81.1}\mmid88.7 & 87.3\mmid\tf{77.3}\mmid88.0 \\
{Yelp} &87.9\mmid\tf{75.4}\mmid86.6 & 86.4\mmid\tf{73.5}\mmid85.0 & 88.3\mmid\tf{76.4}\mmid87.0 \\
\bottomrule
\end{tabular}
}
\caption{Class-wise performance by the finetuned model \tf{Roberta-Large}, \tf{OPT-350M}, and \tf{Flan-T5-Large}. Each entry presents the F1 score of \ti{Positive}/\ti{Entailment}, \ti{Neutral}, and \ti{Negative}/\ti{Contradiction} concatenated with \mmid. \tf{Bold} indicates the lowest F1 score among classes. The results are averaged by 3 runs with different seeds, as shown in Appendix \ref{appendix:hardness:experiment}.}
\label{tab:hardness_model}
\end{table}

\paragraph{Fine-tuning}
To generalize empirical class-wise hardness, models from diverse architectures are chosen:
we use Roberta-Large \cite{DBLP:journals/corr/abs-1907-11692}, OPT-350M \cite{DBLP:journals/corr/abs-2205-01068}, and Flan-T5-Large \cite{chung2024scaling}.
These models belong to encoder-only, decoder-only, and encoder-decoder structures, respectively.
We train these three models separately on eight datasets following the training setups presented in Appendix \ref{appendix:hardness:experiment}.
We select the checkpoint with the best F1 score on the validation dataset to evaluate the test set.
Table \ref{tab:hardness_model} shows that \ti{Neutral} performs poorest among classes with all three models on all the datasets, verifying \ti{Neutral}'s consistent hardness w.r.t. fine-tuned LMs.

\paragraph{In-context Learning}
Beyond model fine-tuning, we also explore another paradigm using large language models (LLMs), where the answer is elicited from LLMs by injecting a cue or instruction \cite{DBLP:journals/corr/abs-2302-14691}.
Specifically, we conduct experiments on MNLI and SNLI using Flan-T5-XXL and LLaMA-2-13B.
The templates employed are shown in Appendix \ref{appendix:hardness:llms}, and \ti{Neutral}'s relative hardness stands referring to Table \ref{tab:appendix:llm}.

\begin{figure}[t]
\centering
\includegraphics[width=0.65\linewidth]{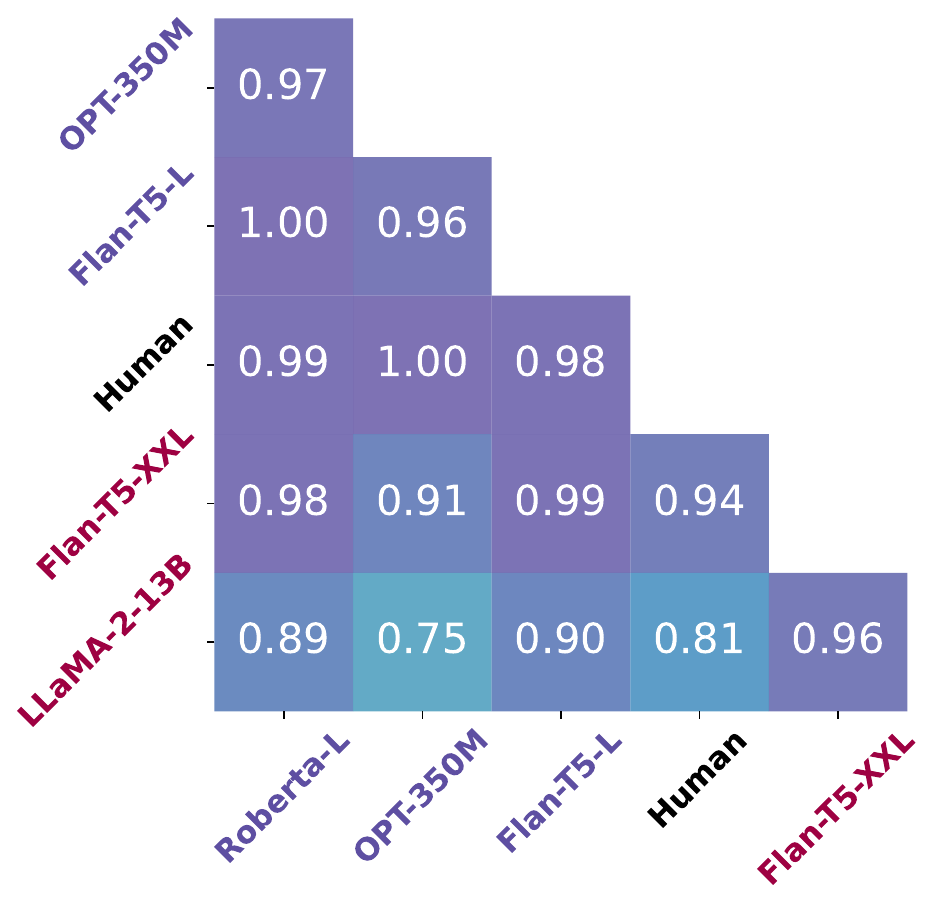}
\vspace{-0.3\baselineskip}
\caption{Correlation matrix among class-wise F1 scores of three {\color{redpolar}finetuned} models together with two {\color{bluepolar}ICL}s and class-wise human disagreement on SNLI, where the high consistency is noted. Figure \ref{fig:mnli_corr} presents MNLI's correlation matrix in Appendix \ref{appendix:hardness:llms}.}
\label{fig:snli_corr}
\end{figure}

Figure \ref{fig:snli_corr} demonstrates \ti{Neutral}'s consistent hardness, e.g., in SNLI, across various LMs $m$, learning paradigms $l$, and human annotation, revealing that its class-wise hardness is \emph{intrinsic}.
Given this observation, we further relax the approximation of $\tilde{\mathrm{H}}$ in Equation \ref{eq:margin} that if the correlation among $\tilde{\mathrm{H}}$ with ($m$, $l$) pairs is higher than a specific threshold, we can approximate $\mathrm{H}$ with $\tilde{\mathrm{H}}$ with arbitrary $m$ and $l$:

\begin{align}
    \mathrm{H}(c_k \mid \mathcal{C}) \approx \tilde{\mathrm{H}}(c_k \mid \mathcal{C}, m, l)
\end{align}

\section{\ti{GeoHard} for class-wise hardness measurement} \label{sec:geohard}
Regarding the intrinsic class-wise hardness shown in Section \ref{sec:hardness}, we quantitatively measure the corresponding empirical hardness motivated by its semantic properties, e.g., \ti{Diverse} and \ti{Middlemost} semantics of \ti{Neutral}.
Specifically, as the name suggests, \ti{GeoHard} measures class-wise hardness by computing the geometrical metrics in the semantic embedding space.

\subsection{Notations} \label{hardness:notations}
The set of $K$ classes is denoted as $\mathcal{C} = \{c_1, ..., c_K\}$. The dataset with $N$ instances is denoted as $\mathcal{D} = \{(X, y)_{1:N}\}$, where $X$ is the input and $y \in \mathcal{C}$ is the corresponding label.
And $\theta$ signifies model parameters.
$\| \cdot \|_{1(2)}$ presents L1(2)-norm.
The input instances of the label $c_k$ is denoted by $X^{c_k}$, i.e., $X^{c_k} = \{X_i | \forall (X_i, y_i) \in D_{train}, y_i = c_k\}$.
\definecolor{color2bg}{HTML}{8C7621}
\subsection{\ti{GeoHard}}\label{geohard_def}
\begin{figure}[t]
		\centering
		\includegraphics[width=0.75\columnwidth]{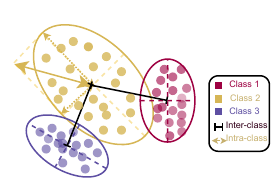}
		\caption{The illustration of \ti{GeoHard} in semantic embeddings space. The ellipses approximate class-wise data distribution. {\color{color2bg}Class 2} is speculated to be difficult due to its large variance and middlemost location.}
		\label{fig:metrics_illustration}
\end{figure}

\paragraph{Semantic representation}
As \ti{GeoHard} aims to measure class-wise hardness through modeling semantics, a sentence encoder is therefore required, which maps a sentence to a vector with a dimension $E$.
We denote this mapping function as $f(\cdot)$.

\paragraph{Semantics-guided metrics}
\ti{GeoHard} consists of \ti{intra-} and \ti{inter-}class metrics modeling two semantics properties, as illustrated in Figure \ref{fig:metrics_illustration}.
The \ti{intra}-class metric, corresponding to \ti{Diverse} semantics, quantifies the distributional variance within one class, formulated as:
\begin{align}
    \mathrm{H}_{intra}(c_k) = \| \sigma(f(X^{c_k})) \|_2
    \label{eq:intra}
\end{align}

where $\sigma$ denotes the element-wise variance across the instances, i.e., $\sigma : \mathbb{R}^{N \times E} \rightarrow \mathbb{R}^E$.

\ti{Middlemost} semantics indicate one class is located closer to other classes in the representation space.
Hence, the \ti{inter}-class metric calculates the average distance from one class center to the other classes.
The opposition aims to unite $m_{inter}$ with the $m_{intra}$ regarding the overall hardness tendency:
\begin{align*}
    \mathrm{H}_{inter}(c_k) & = \frac{-\sum^{K}_{\substack{i=1\\ i \neq k}} ||\mu(f(X^{c_k})) - \mu(f(X^{c_i}))||_1}{K-1}
\end{align*}

where $\mu(\cdot)$ presents the element-wise mean operation across the input set, that is $\mu: \mathbb{R}^{N \times E} \rightarrow \mathbb{R}^E$.

To this end, \ti{GeoHard} of one specific class is the amalgamation of the class-wise \ti{intra-} and \ti{inter-class} metrics, i.e., $\mathrm{H}_{GeoHard}(c_k) = \mathrm{H}_{intra}(c_k) + \mathrm{H}_{inter}(c_k)$. And the higher \ti{GeoHard} indicates more class-wise hardness, i.e., a poorer performance.

\subsection{Implementation}
According to the open reference \footnote[4]{E5-large-v2 led \hyperlink{https://huggingface.co/spaces/mteb/leaderboard}{Massive Text Embeddings Benchmark} leaderboard \cite{DBLP:conf/eacl/MuennighoffTMR23} at the time of the work.}, we apply E5-large-v2 \cite{DBLP:journals/corr/abs-2212-03533} to project sentences to a high dimensional space.
\citet{jha-mihata-2021-geodesic} point out that the nonlinear dimension reduction on contextualized representation benefits downstream tasks.
Therefore, we apply Uniform Manifold Approximation and Projection (UMAP; \citealt{McInnes2018}) to compress sentence representation to $E$ \footnote[5]{We set $E=2$ for visualization in the experiment.}.
The complete encoder consists of E5-large-v2 and UMAP \footnote[6]{As E5-large-v2 is trained to capture uni-sentence semantics, we concatenate premise and hypothesis in NLI tasks with six conjunctive words or phrases shown in Appendix \ref{appendix:our_eval:connecting} referring to the templates applied in \citealt{gao2021making}.}.

\definecolor{mycolor}{RGB}{255, 204, 204}

\begin{table*}[t]
\centering
\small
\begin{tabular}{l|ccccc|ccc|c}
\toprule
{\multirow{2}{*}{\backslashbox{\makebox[1em][l]{Metric}}{\makebox[2.4em][l]{Dataset}}}} & \multicolumn{5}{c|}{\tf{SC}} & \multicolumn{3}{c|}{\tf{NLI}} & \multirow{2}{*}{\shortstack{Macro Avg. $\uparrow$ \\ (Absolute)}} \\
&\tf{Amazon} & \tf{APP} & \tf{SST-5} & \textbf{TFNS} & \textbf{Yelp} & \textbf{MNLI}  &\textbf{SNLI} & \textbf{SICK-E} &  \\
\midrule
SA &  .4730&\colorbox{mycolor}{-.9620}&\colorbox{mycolor}{-.2244}&\colorbox{mycolor}{-.9047}&.8980&\colorbox{mycolor}{-.7219}&.3930&\colorbox{mycolor}{-.9962}&.2556\tiny{$\pm$.4961} \\ 
\ti{Thrust} &\colorbox{mycolor}{-.9780}&.9012&.0833&.5157&.9952& .0000 &.8311& .0000 &.2936\tiny{$\pm$.3792} \\
Spread  &\colorbox{mycolor}{-.6350}&.4550&.8369&.9944&\colorbox{mycolor}{-.2148}&.4292&\colorbox{mycolor}{-.3934}&.8471&.2899\tiny{$\pm$.3412} \\
\midrule
\ti{GeoHard-Intra}  &.5857&.9539&.9892&\colorbox{mycolor}{-.1574}&.9947&.9784&.8805&.6647&.7362\tiny{$\pm$.1354} \\
\ti{GeoHard-Inter} &.9997&.9964&.9908&.9722&.9978&.1500&.9042&.3663&.7972\tiny{$\pm$.1006} \\
\ti{GeoHard} &.9998&.9958&.9909&.8852&.9977&.8384&.8882&.4871&\tf{.8854}\tiny{$\pm$\tf{.0262}} \\
\bottomrule
\end{tabular}
\caption{Pearson's correlation coefficients between class-wise hardness measurement and class-wise F1 scores, i.e., the approximation of $\mathrm{H}$.
All the metrics have been adjusted so that the higher correlation indicates better measurement.
\colorbox{mycolor}{Red} indicates that the value is opponent to the original design.
\tf{Bold} indicates the best performance among the methods with the highest average correlation and lowest variance across the datasets.
$\uparrow$ indicates the higher values present better results.
\ti{GeoHard}'s results are averaged on 3 runs (and 6 conjunctions for NLI).
Please refer to Appendix \ref{appendix:measurement:result} for the detailed values. }
\label{tab:corr_table}
\end{table*}

\section{Experiments} \label{sec:exp}
\subsection{Baseline: \ti{Instance} Hardness Aggregation}
\paragraph{Sensitivity Analysis \cite{DBLP:journals/tacl/HahnJF21}}
measures data hardness by assessing how perturbations in the input affect a model's prediction.
It calculates the model's prediction confidence for an instance and its perturbed neighbor on the golden label.
A larger derivative between these confidences, i.e., higher sensitivity, signifies greater hardness.

As for the class-wise hardness, we average the sensitivity values of the samples in each class.
The higher class-wise sensitivity suggests more difficulty in the class, in consistency with \citet{DBLP:journals/tacl/HahnJF21}. 
We take the finetuned Roberta-Large in Section \ref{sec:hardness_LMs} as the reference model.
More implementation details can be found in Appendix \ref{appendix:measurement:sa}.

\paragraph{Spread \cite{DBLP:conf/emnlp/ZhaoMM22} \& Thrust \cite{zhao2023thrust}}
measure the instance-level hardness by estimating the similarity between test instances and training samples.
Concretely, Spread calculates the semantic similarity between test instances and a few-shot closest training samples, using the sentence encoders.
E5-large-v2 \cite{DBLP:journals/corr/abs-2212-03533} is also applied by Spread in line with \ti{GeoHard}, and the number of training selections is 8.
Thrust calculates the distance of the decoded instance representation by LLMs between training and test sets.
We apply the identical LLM as the original work, i.e., Flan-T5-Large fine-tuned on UnifiedQA dataset \footnote[7]{https://huggingface.co/allenai/unifiedqa-t5-large} \cite{DBLP:conf/emnlp/KhashabiMKSTCH20}.

As both methods are similarity-based, the smaller similarity indicates more hardness.
To this end, we average Spread scores in each class as the class-wise metrics.
For Thrust, we select the bottom 25 percentile of Thrust scores in each class as the aggregation \footnote[8]{The reason we do not average Thrust for class-wise hardness here is that this metric is inversely proportional to the distance. Therefore, Thrust values will come to infinity when the test sample is extremely close to the training set.}.
Appendix \ref{appendix:measurement:thrust} and \ref{appendix:measurement:spread} present their detailed implementation.

\subsection{Quantification of class-wise hardness}
We benchmark the instance-aggregating methods (SA, Spread, and Thrust) as well as \ti{GeoHard}, including its \ti{intra-} and \ti{inter-}class metrics, on the eight NLI/SC datasets in Section \ref{sec:hardness}.

Section \ref{sec:hardness} illustrates the consistency between LMs and humans regarding class-wise hardness, and this allows us to select an arbitrary empirical class-wise hardness $\tilde{\mathrm{H}}$ as a \ti{close} approximation of $\mathrm{H}$.
Consequently, we apply the class-wise F1 scores from fine-tuned Roberta-Large as the hardness reference.
Following the previous work \cite{DBLP:conf/emnlp/ZhaoMM22}, we determine the effectiveness of various metrics by calculating \ti{Pearson}'s correlation coefficient between metrics and the hardness reference (Table \ref{tab:corr_table}).
Considering negative correlation, we take the \tf{absolute} value of average correlations as shown in the right-most column, namely that higher values indicate better measurement.

\subsection{Analysis on Experimental Results}
Table \ref{tab:corr_table} presents the correlation between class-wise hardness measurement and the reference hardness on these eight NLI/SC datasets.
The class-wise SA, Spread, and Thrust are shown to be \tf{poorly} correlated to the reference, with average correlations of 0.2556, 0.2936, and 0.2899, respectively.
Their large variance in correlation across tasks indicates their incompetence in class-wise hardness measurement.
Meanwhile, \ti{GeoHard} significantly outperforms these instance-level methods, exhibiting the lowest variance across the tasks.
In addition to these metrics, \ti{GeoHard} surpasses its components, namely the \ti{intra-} and \ti{inter-}class metrics, highlighting their complementarity and underscoring \ti{GeoHard}'s comprehension of class-specific properties.

\section{Generalization of \ti{GeoHard}}
\label{sec:generalization}
The previous section showcased the exceptional performance of \ti{GeoHard} in measuring hardness in NLI and SC tasks by leveraging class-wise semantic properties.
In this section, we explore \ti{GeoHard}'s robustness and generalization both theoretically and empirically.
We conduct the experiments to demonstrate \ti{GeoHard}'s generalization capabilities across various sentence encoders and other types of tasks, further substantiating the connection between class-wise hardness and semantics.
Furthermore, we highlight \ti{GeoHard}'s robustness in low resource scenarios, showcasing its advantage as a training-free metric.

\subsection{Theoretical proof on generalization}
\begin{figure}[t]
\centering
\includegraphics[width=0.75\linewidth]{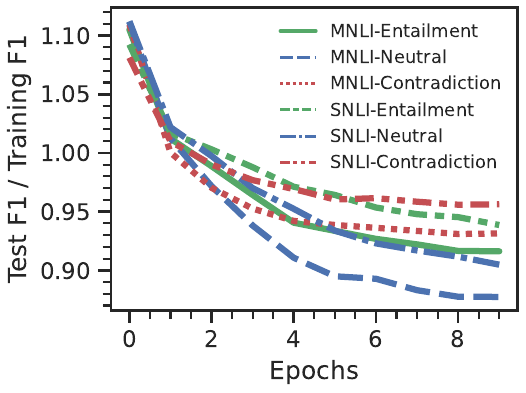}
\caption{The ratio between F1 scores on the test and training sets for training epochs on NLI tasks. \ti{Neutral} \blue{in blue} suffers from overfitting most. Figure \ref{fig:sc_dist_diff} in Appendix \ref{appendix:general:figure} presents a similar issue in NLI tasks.}
\label{fig:dist_diff}
\end{figure}

\textit{GeoHard}'s robustness is evident in its ability to effectively elucidate factors contributing to class-wise hardness, such as overfitting depicted in Figure \ref{fig:dist_diff}. The \textit{intra-class} metric within \textit{GeoHard} serves to gauge the extent of overfitting, namely, the divergence between training and test data, as elaborated in the following Theorem \ref{thm1}.

\begin{theorem}
Assuming a Gaussian distribution for instances within $c_k$, $D \sim \mathcal{N}(\mu_{c_k}, \sigma_{c_k}^2)$, the means of the training and test data can be represented as $\hat{\mu}_{c_k}^{tr} \sim \mathcal{N}(\mu_{c_k}, \sigma_{c_k}^2 / n_{tr})$ and $\hat{\mu}_{c_k}^{te} \sim \mathcal{N}(\mu_{c_k}, \sigma_{c_k}^2 / n_{te})$, where $n_{tr}$ and $n_{te}$ are the sizes of the training and test sets within $c_k$, respectively. Note that, conditioned on the class $c_k$, the data $D$ is i.i.d. as mentioned above. By applying Chebyshev's inequality \cite{mitrinovic2013classical}, the following inequality holds for any arbitrary $k \in \mathbb{R}_{+}$ (see mathematical derivation in Appendix \ref{appendix:general:proof}):
\begin{align}
 \frac{2}{k^2} \geq P\left(|\hat{\mu}_{c_k}^{tr} - \hat{\mu}_{c_k}^{te}| \geq \frac{2k \sigma_{c_k}}{ \sqrt{ n_{te}}}\right)
\end{align}
\label{thm1}
\end{theorem}

Hence, for any arbitrary $k$, the data variance $\sigma_{c_k}$ reflected by $\mathrm{H}_{intra}(c_k)$, as depicted in Equation \ref{eq:intra}, serves as an estimation for the distributional gap $|\hat{\mu}_{c_k}^{tr} - \hat{\mu}_{c_k}^{te}|$, indicating the overfitting degree of $c_k$.

\subsection{Cross-embeddings generalization}
\begin{figure}[t]
\centering
\includegraphics[width=0.65\linewidth]{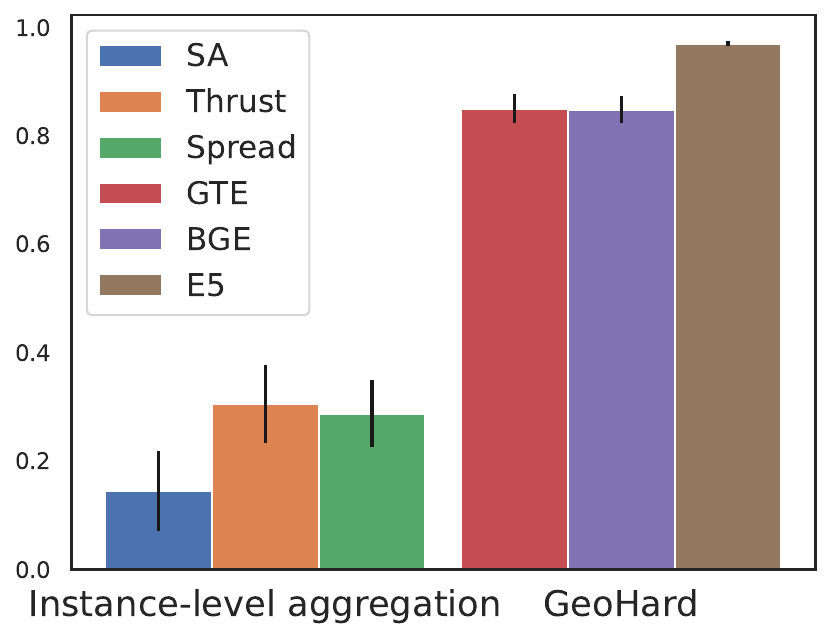}
\caption{Average \ti{Pearson}'s coefficient between various metrics and hardness reference on five SC tasks. \ti{GeoHard} with different embeddings consistently and significantly outperform instance-level aggregation, demonstrating the robustness of \ti{GeoHard}.}
\label{fig:ins_vs_geohard}
\end{figure}

We incorporate two other architectures of sentence embeddings, i.e., GTE-large \cite{DBLP:journals/corr/abs-2308-03281} and BGE-large-en-v1.5 \cite{DBLP:journals/corr/abs-2309-07597} into \ti{GeoHard}, substituting E5-large-v2.
Observing Figure \ref{fig:ins_vs_geohard}, we find a consistent trend of \ti{GeoHard}'s measurement across different sentence embeddings.
The significant gap between \ti{GeoHard} and the instance-level aggregation underscores the robustness of \ti{GeoHard} as a semantic-guided metric.

\subsection{Cross-task generalization}
Complementary to the theoretical generalization, we further validate \ti{GeoHard} on other tasks beyond SC and NLI, i.e., topic classification and emotion detection.
We include AG News \cite{DBLP:conf/nips/ZhangZL15}, Yahoo Answer Topic (Yahoo; \citealt{DBLP:conf/nips/ZhangZL15}) for the former and Emo2019 (Emo; \citealt{DBLP:conf/semeval/ChatterjeeNJA19}), Contextualized Affect Affect Representations for Emotion Recognition (CARAR; \citealt{DBLP:conf/emnlp/SaraviaLHWC18}) for the latter.

We fine-tune Roberta-Large on these four datasets to obtain the reference empirical hardness, i.e., class-wise F1 scores, and also conduct \ti{GeoHard}, referring to Table \ref{tab:appendix:agnews}-\ref{tab:appendix:yahoo} in Appendix \ref{appendix:general:empirical_proof}.
According to Table \ref{tab:corr_others}, the consistency between the measurement and reference on the tasks other than NLI and SC empirically exhibits the generalization of \ti{GeoHard} for class-wise hardness measurement.

\begin{table}[t]
\centering
\small
\begin{tabular}{l|cccc}
\toprule
 & \tf{AG News} & \tf{Yahoo} & \tf{Emo} & \tf{CARAR}  \\
\midrule
\ti{GeoHard}   & 
-.980\tiny{$\pm$.0}&-.838\tiny{$\pm$.0}&-.798\tiny{$\pm$.1}&-.817\tiny{$\pm$.1}\\
\bottomrule
\end{tabular}
\caption{Pearson's correlation coefficients, averaged on 3 seeds, between class-wise hardness measured by \ti{GeoHard} and class-wise F1 scores, i.e., the reference of hardness, on topic classification and emotion detection.}
\label{tab:corr_others}
\end{table}

\subsection{Robustness in low-resource scenarios}
\begin{figure}[t]
\centering
\includegraphics[width=\linewidth]{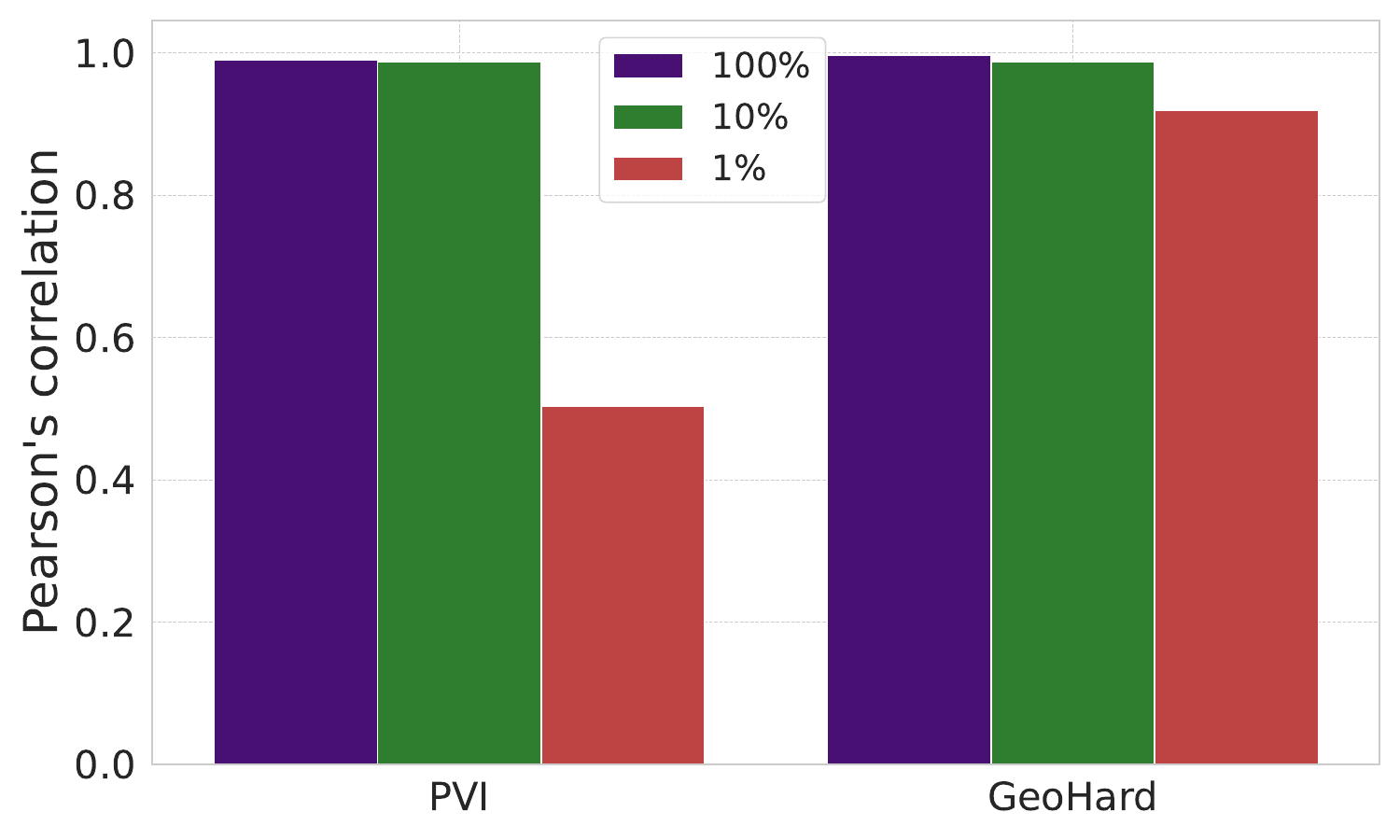}
\caption{Performance comparison of \textit{GeoHard} and \texttt{PVI} in low-resource scenarios: \textit{GeoHard} experiences less degradation on the average Pearson's correlation (absolute values) across five SC datasets with 1\% of the training data compared with the full training data.}
\label{fig:low_resource}
\end{figure}

In this section, we will demonstrate the robustness of our method in low-resource scenarios.
 We have randomly selected 1\%, 10\%, and 100\% of the instances from the training corpus across five SC datasets included in Section \ref{sec:exp}.
As illustrated in Figure \ref{fig:low_resource}, \textit{GeoHard} exhibits notably less performance degradation in low-resource settings compared to \texttt{PVI} \cite{DBLP:conf/icml/EthayarajhCS22}, underscoring its robustness as a training-free method.

\section{Why class-wise hardness measurement?}
\label{sec:application}
In the previous sections, we establish the concept of class-wise hardness, which can be well and robustly measured by \ti{GeoHard}.
One of the most relevant literature for the application of \ti{GeoHard} is class reorganization.

Class reorganization has received relatively limited attention compared to research focusing on addressing class imbalance \cite{DBLP:conf/emnlp/Subramanian0BCF21, DBLP:conf/eacl/HenningBFF23}, primarily due to extra annotation.
However, initial task formulations are rarely perfect, and as research progresses, class reorganization becomes necessary for a more comprehensive understanding and effective modeling of the task.
For example, in NLI, the task evolved from a 2-way classification \cite{DBLP:conf/mlcw/DaganGM05} to a 3-way classification by separating \ti{Non-Entailment} into \ti{Neutral} and \ti{Contradiction}.
Recently, \citet{nighojkar-etal-2023-strong} further subdivided \ti{Neutral} into two distinct classes based on human disagreement.

As for the practical perspective, class reorganization can balance the model performance among the classes \cite{DBLP:conf/acl/PottsWGK20}.
To resolve the severe imbalance between \ti{Neutral} and other classes in SC dataset, as shown in Table \ref{tab:hardness_model}, Dynasent \cite{DBLP:conf/acl/PottsWGK20} opted to split \ti{Neutral} to \ti{Mixed} (a mixture of positive and negative sentiment) and \ti{Neutral} (conveying nothing regarding sentiment).
This approach aims to achieve a \tf{coherent} categorization, which narrows the performance gap among classes.

In this section, we use Dynasant \cite{DBLP:conf/acl/PottsWGK20} as an example to demonstrate how measuring class-wise hardness can aid in interpreting class reorganization and facilitate the learning process.

\subsection{\ti{GeoHard} interprets class reorganization}
\ti{GeoHard} can provide insights into two crucial questions regarding class reorganization: \ti{what} and \ti{how} to reorganize classes.
Firstly, even without training a model, \ti{GeoHard} can directly provide hardness estimates across the original classes to locate the operating target.
Secondly, \ti{GeoHard} can assess the effectiveness of the formulation strategy and hence guide the class reorganization.

We conduct \ti{GeoHard} on classes before and after Dynasent's class reorganization, which splits \ti{Neutral} to \ti{Mixed} and \ti{Neutral}.
For comparison, we randomly split \ti{Neutral} into two classes, labeled \ti{Rand1} and \ti{Rand2}.
As illustrated in Figure \ref{fig:dynasent_geo_illu}, the \ti{Neutral} and \ti{Mixed} are shown to be highly separable, indicating their distinction in semantics.

As shown in Table \ref{tab:dynasent}, the new classes \ti{Mixed} and \ti{Neutral} exhibit lower class-wise hardness compared to the original \ti{Neutral}.
The standard derivation among class-wise hardness on the newly organized Dynasent drops by 40.9\% (from 1.15 to 0.68).
This clearly explains the \tf{coherent} class formation by reorganizing \ti{Neutral} into two sub-classes.
However, not all class reorganizations yield beneficial outcomes: a random split may result in high overall class-wise hardness and a severe imbalance of class-wise hardness.

\definecolor{darkyellow}{RGB}{180,130,0}
\definecolor{darkgreen}{RGB}{0,100,0}
\definecolor{biege}{RGB}{254, 255, 184}

\begin{figure}[t]
    \centering
    \begin{subfigure}[b]{0.23\textwidth}
        \centering
        \includegraphics[width=1.25\textwidth]{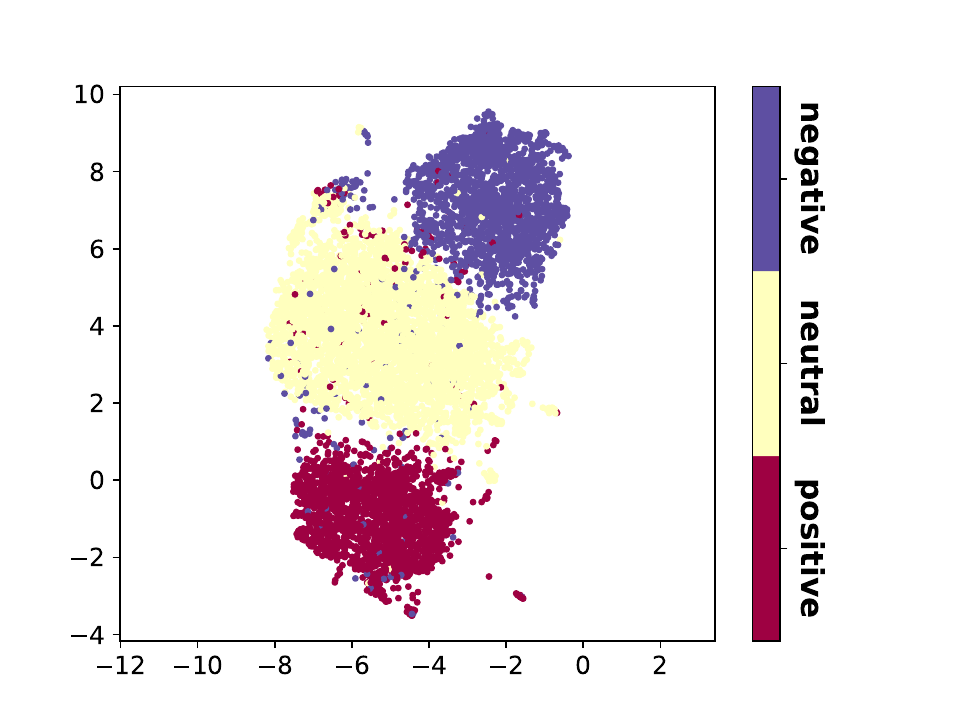}
        \caption[Network2]%
        {{\small 3-way}}    
        \label{fig:amazon_geo}
    \end{subfigure}
    \hfill
    \begin{subfigure}[b]{0.23\textwidth}  
        \centering 
        \includegraphics[width=1.25\textwidth]{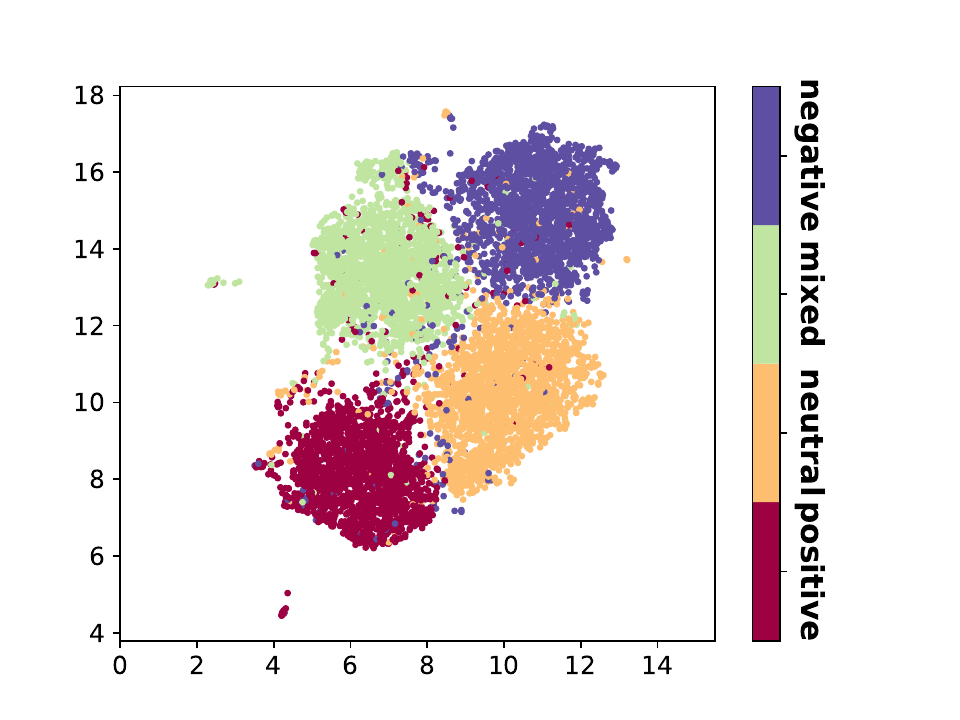}
        \caption[]%
        {{\small 4-way}}    
        \label{fig:sst5_geo}
    \end{subfigure}
    \caption{Illustration of the class-wise geometrical distribution of 3-way and 4-way Dynasent by splitting \textcolor{darkyellow}{\ti{Neutral}} to \textcolor{darkgreen}{\ti{Mixed}} and \textcolor{orange}{\ti{Neutral}} while maintaining \textcolor{red}{\ti{Positive}} and \textcolor{blue}{\ti{Negative}}. The reorganized class \textcolor{darkgreen}{\ti{Mixed}} and \textcolor{orange}{\ti{Neutral}} are highly separable in the representation space.}
    \label{fig:dynasent_geo_illu}
\end{figure}

\begin{table}[t]
\centering
\small
\begin{tabular}{c|cc|c|c}
\toprule
\ti{\tf{Positive}} & \multicolumn{2}{c|}{\ti{\tf{Neutral}}} & \ti{\tf{Negative}} & \tf{Std} \\
-5.58  & \multicolumn{2}{c|}{-3.08} & -5.48   & 1.15  \\ \midrule 
\ti{\tf{Positive}} & \ti{\tf{Rand1}}   & \ti{\tf{Rand2}} & \ti{\tf{Negative}} & \tf{Std} \\ 
-3.89             & -0.71            & -0.71          & -3.62            & 1.53  \\ \midrule
\ti{\tf{Positive}} & \ti{\tf{Mixed}} & \ti{\tf{Neutral}} & \ti{\tf{Negative}} & \tf{Std} \\
-5.24             & \underline{-4.26}   &  -3.41          & -4.78            & \tf{0.68} \\ \bottomrule
\end{tabular}
\caption{Class-wise \ti{GeoHard} on the 3-way and reorganized Dynasent (randomly and semantic-guided splits). The results are averaged on 3 seeds. Note that larger \ti{GeoHard} indicates more hardness on the class. \tf{Bold} signifies the smallest standard derivation among class hardness. \underline{Underline} indicates the lowest hardness on \ti{Neutral} and its splits.}
\label{tab:dynasent}
\end{table}

\subsection{\ti{GeoHard} propels task learning}
We have demonstrated that \ti{GeoHard} can interpret and validate the reorganization of labels. Next, we further investigate how to leverage the class-wise hardness knowledge and its induced class reorganization to enhance task learning, with methods such as ICL.
Typically, ICL samples the demonstrations uniformly across classes \cite{DBLP:conf/emnlp/MinLHALHZ22}.
Here, we demonstrate the benefits of splitting the \tf{hardest} class into two \tf{easier} ones in ICL, elucidating the significance of class-wise hardness.

We divide each class into two sub-classes and select instances from these newly formed classes.
For classes \ti{Positive} and \ti{Negative}, which lack prepared sub-classes, we employ KMeans on the embeddings to separate instances within each class into two sub-classes \cite{zhang2023automatic,yang-etal-2023-representative}.
Then, we select the center of each cluster as a demonstration.
For example, if \ti{Positive} is selected for reorganization, the demonstrations consist of 2 \ti{Positive} instances, 1 \ti{Neutral} instance, and 1 \ti{Negative} instance.
For convenience, we abbreviate the selection as \textsc{2P+1Neu+1N}.

We randomly sample 1,000 instances from each class from 3-way Dynasent \cite{DBLP:conf/acl/PottsWGK20}, wherein \ti{Neutral} class contains 500 \ti{Mixed} instances and 500 new-formed \ti{Neutral} instances. We conduct ICL on two popular LLMs, i.e., OPT-6.7B \cite{DBLP:journals/corr/abs-2205-01068}  and Llama-2-7B-32K-Instruct \cite{DBLP:journals/corr/abs-2307-09288} with different setups of demonstrations:
(1) even sampling: \textsc{1P+1Neu+1N}; (2) sampling based on class reorganization: \textsc{2P+1Neu+1N}, \textsc{1P+2Neu+1N}, and \textsc{1P+1Neu+2N}. For both setups, we select the centroid instance from each cluster.
The examples of demonstrations can be found in Appendix \ref{appendix:general:application_demostrations}.

As shown in Table \ref{tab:dynasent_icl}, both models utilizing the setup \textsc{1P+2Neu+1N} attain the best performance, which is the advocated action from \ti{GeoHard} since \ti{Neutral} is measured as the hardest class and the new classes \ti{Mixed} and \ti{Neutral} are relatively easier.
However, reconstructing other classes may not lead to such benefits in learning and can even lead to significant degradation (e.g., \textsc{2P} with LLama-7B).

\begin{table}[t]
\centering
\small
\begin{tabular}{c|c|c}
\toprule
Demonstration \ \% & OPT-6.7B & LLama-7B \\
\midrule
\textsc{1P+1Neu+1N} & 61.7\tiny{$\pm$3.54} & 61.4\tiny{$\pm$1.82} \\
\textsc{2P+1Neu+1N} & 61.1\tiny{$\pm$0.02} & 39.6\tiny{$\pm$10.71}\\
\textsc{1P+2Neu+1N} & \tf{64.3}\tiny{$\pm$1.24} & \tf{69.9}\tiny{$\pm$1.74} \\
\textsc{1P+1Neu+2N} & 60.4\tiny{$\pm$1.69} & 34.9\tiny{$\pm$3.36} \\
\bottomrule
\end{tabular}
\caption{Comparison of different compositions of demonstrations on Dynasent, with each entry presenting the prediction accuracy. \tf{Bold} indicates the highest accuracy on one specific model. The results are averaged on three seeds for random initialization in KMean.}
\label{tab:dynasent_icl}
\end{table}

\section{Related works}
\paragraph{Hardness in NLP datasets}
Instance-level hardness indicates the difficulty of an instance given a distribution \cite{DBLP:conf/icml/EthayarajhCS22}, and the taxonomy is summarized in Figure \ref{fig:tax} in Appendix \ref{appendix:measurement:taxonomy}.
Without training, the reference model or embedding is usually needed.
With a model as the reference, Sensitivity Analysis \cite{DBLP:journals/tacl/HahnJF21,chen-etal-2023-relation} assesses hardness by perturbing input features and observing the resulting changes in model predictions.
Additionally, Thrust \cite{DBLP:journals/corr/abs-2307-10442} approximates instance hardness based on the external knowledge required by an LLM.
In parallel with the model reference, Spread \cite{DBLP:conf/emnlp/ZhaoMM22} leverages the similarity between test and training samples in the space of semantic embeddings for hardness measurement.
Alternatively, information theory-based methods, such as point-wise \textcal{V}-usable information (\ts{PVI}; \citealt{DBLP:conf/icml/EthayarajhCS22}) and Rissanen Data Analysis (RDA; \citealt{DBLP:conf/icml/PerezKC21}), offer insights into data hardness using training outcomes.
Moreover, other methods measure data hardness from training dynamics, including dataset cartography \cite{swayamdipta-etal-2020-dataset}, forgetting scores \cite{toneva2018an}, and Error L2-Norm \cite{paul2021deep}, etc \footnote[9]{We include the class-wise hardness measurement with some training-based methods and training-dynamics methods in Appendix \ref{appendix:other_metrics}}.
This work primarily focuses on the training-free methods, which are more practical and scalable for gauging hardness with LLMs.

Although instance-level hardness is well studied, class-wise hardness is under-explored.
Therefore, our work explores the class-wise measurement by aggregating the existing instance-level methods first and then specifically designs \ti{GeoHard}, which requires no additional data or training.

\paragraph{Geometrical view of classification complexity}
In the context of general machine learning, prior research \cite{ho2002complexity, lorena2019complex} assesses the difficulty of a classification problem through the analysis of data geometry and inter- and intra-class distribution.
Various metrics of quantification, such as Fisher's discriminant ratio \cite{DBLP:phd/basesearch/Cummins13}, overlapping regions \cite{seijo2019developing} and network measures \cite{DBLP:journals/ijon/GarciaCL15}, have been proposed to qualify class-wise complexity based on geometric features.

Sentences encoders \cite{DBLP:conf/emnlp/ReimersG19} excel at generating high-dimension sentence embeddings based on semantics.
We explore class-wise hardness by leveraging geometrical features within and among the classes, inspired by \ti{Neutral}'s specific semantics.

\paragraph{\ti{Neutral} in NLU}
\ti{Neutral} depicts undetermined or middlemost semantics while ruling out other classes, and widely exists in NLU tasks such as NLI \cite{DBLP:conf/naacl/WilliamsNB18, DBLP:conf/emnlp/BowmanAPM15} and SC \cite{DBLP:conf/iclr/SunDNT19}.
Generally, the class with the prefix \ti{Non-} also delivers similar semantics with \ti{Neutral}, i.e., excluding other classes.
For instance, the Microsoft Research Paraphrase (MRPC; \citealt{DBLP:conf/acl-iwp/DolanB05}) dataset aims to determine whether a pair of questions are semantically equivalent, i.e., to classify sentence pairs to \ti{Equivalent} and \ti{\tf{Non-}equivalent}.
In GLUE \cite{DBLP:conf/iclr/WangSMHLB19}), six of nine tasks contain a \ti{Neutral} or \ti{Non-} class, indicating the wide existence of classes with undetermined semantics in NLU.

Due to \ti{Neutral}'s semantic prevalence, we initiate class-wise hardness from exploring the tasks containing \ti{Neutral} and then extend to general classes.

\section{Conclusion}
In this work, to study how class-specific properties influence model learning, we initiate the notion of class-wise hardness analogous to instance-level hardness.
The consistent pattern observed across various LMs, learning paradigms, and human annotations on eight NLU datasets affirms the presence of class-wise hardness as an inherent property.
In addressing the challenge of estimating class-wise hardness, conventional instance-level metrics fall short, necessitating a tailored approach to measure hardness specifically at the class level.
Thus, we introduce \textit{GeoHard}, which models both \ti{inter-} and \ti{intra-}class semantics, surpassing instance-level aggregation by 59\%.
Moreover, \ti{GeoHard}, formulated with a foundation in semantics, demonstrates robust generalization properties, as validated both theoretically and empirically.
Lastly, we showcase the potential of \ti{GeoHard} in reorganizing classes and enhancing task-learning methodologies.
We recommend more attention to class-wise hardness and exploring its potential across a broader range of scenarios.

\section*{Limitations} 
Our work, introducing the concept of class-wise hardness and proposing a practical metric, does come with specific limitations that justify further exploration.
Firstly, as an initiative work, we only cover limited types of classification tasks in NLU due to the space constraint.
Some classification problems such as sequence labeling \cite{DBLP:journals/corr/abs-2011-06727} are not covered in our scope.
Class-wise hardness for other formats of classification tasks is still obscure and needs further exploration.
Secondly, as our proposed method \ti{GeoHard} is built upon the pre-trained sentence encoders, they inherit their corresponding limitations.
For example, it will be intricate to measure class-wise hardness in the complex semantics or long inputs.
These cases can not be well modeled by a single-sentence encoder yet.
Combining the two problems above leads to a new issue.
Hypothetically, given the assumption that the class-wise hardness for other formats of NLP problems still exists, how to model them will be the potential concern, as it is beyond the capacity of sentence encoders.
Regarding the application of \ti{GeoHard} and the class-wise hardness that it measures, we have not gone deeper into this problem.
A larger-scale study is expected to further explore this topic.
In conclusion, further efforts are expected to overcome the limitations of this work.

\section*{Ethical Statements} 
We foresee no major ethical concerns in our work.
The datasets we used in this work are all publicly available.
As far as we see, there is no sensitive information included.
For the language models we applied, the outputs, i.e., the class labels, are not sensitive either.

\section*{Acknowledgement}
We thank Sheng Lu, Kexin Wang, Indraneil Paul, Hendrik Schuff, and Sherry Tongshuang Wu for their feedback on an early draft of this work.
Fengyu Cai is funded by the German Federal Ministry of Education and Research and the Hessian Ministry of Higher Education, Research, Science, and the Arts within their joint support of the National Research Center for Applied Cybersecurity ATHENE.
Xinran Zhao is funded by the ONR Award N000142312840.

\bibliographystyle{acl_natbib}
\bibliography{custom}

\clearpage
\onecolumn
\appendix
\part*{Appendix}

\section{Validation of class-wise hardness}

\subsection{Dataset} \label{appendix:data_examples}

\newcolumntype{b}{X}
\newcolumntype{m}{>{\hsize=.9\hsize}X}
\newcolumntype{s}{>{\hsize=.5\hsize}X}

\begin{table*}[htbp]
\small
\centering
\resizebox{\textwidth}{!}{
\begin{tabularx}{\textwidth}{smss}
\toprule
    Dataset & Description & Statistics (train/dev/test)\\
    \midrule
    Amazon Review Multi en $^\dag$ (Amazon; \citealt{DBLP:conf/emnlp/KeungLSS20}) & an Amazon product reviews dataset for multilingual text classification (we only use English part) & 120,000 / 3,000 / 3,000 \\
    \midrule
    App Reviews $^\dag$ (APP; \citealt{DBLP:conf/sigsoft/GranoSMVCP17})        & Android app reviews categorized classifying types of user feedback from a software maintenance and evolution perspective & 56,151 / 6,804 / 6,633 \\
    \midrule
    MultiNLI $^\ddag$ (MNLI; \citealt{DBLP:conf/naacl/WilliamsNB18})     &     Multi-Genre Natural Language Inference annotated with textual entailment information & 353,408 / 39,270 / 9,369\\
    \midrule
    SICK-E $^\ddag$ \cite{DBLP:conf/lrec/MarelliMBBBZ14}   & A dataset targeting Natural Language Inference & 1,920 / 213 / 2,136 \\
    \midrule 
    SNLI $^\ddag$ \cite{DBLP:conf/emnlp/BowmanAPM15}       & Stanford Natural Language Inference Corpus & 548,292 / 9,705 / 9,657 \\
    \midrule 
    SST-5 $^\dag$ \cite{DBLP:conf/emnlp/SocherPWCMNP13}       & Stanford Sentiment Treebank with 5 labels & 4,872 / 1,332 / 1,332 \\
    \midrule
    Twitter Financial News Sentiment $^\dag$ \footnotemark[10] (TFNS) & A dataset is used to classify finance-related tweets for their sentiment & 3,891 / 435 / 1,041  \\
    \midrule
    Yelp review $^\dag$ (Yelp; \citealt{DBLP:conf/nips/ZhangZL15}) & A dataset containing custom reviews from Yelp & 351,000 / 39,000 / 30,000 \\
    \bottomrule
\end{tabularx}
}
\caption{The description of the datasets used in the class-wise hardness measurement, together with the statistics of newly-formulated datasets after balancing the number of instances in each class. All the datasets consist of 3 classes, namely \ti{Positive}/\ti{Neutral}/\ti{Neagtive} in SC $^\dag$ and \ti{Entailment}/\ti{Neutral}/\ti{Contradiction} in NLI $^\ddag$.}
\label{tab:dataset}
\end{table*}

{\footnotetext[10]{\hyperlink{https://huggingface.co/datasets/zeroshot/twitter-financial-news-sentiment}{https://huggingface.co/datasets/zeroshot/twitter-financial-news-sentiment}}}

\begin{table}[h]
\small
\centering
\begin{tabularx}{\textwidth}{smsss}
    \toprule
    Dataset & Example & Original labels & Original statistics\\
    \midrule
    Amazon Review Multi en (Amazon; \citealt{DBLP:conf/emnlp/KeungLSS20}) & Title: bubble  \newline Body: went through 3 in one day doesn't fit correct and couldn't get bubbles out (better without) & 1, 2, 3, 4, 5 & 200,000 / 5,000 / 5,000 \\
    \midrule
    App Reviews (APP; \citealt{DBLP:conf/sigsoft/GranoSMVCP17}) & simple and perfect About this software rtl sdr is very useful ... installed done. Thanks. & 1, 2, 3, 4, 5 & 230,452 / 28,806 / 28,807 \\
    \midrule
    MultiNLI (MNLI; \citealt{DBLP:conf/naacl/WilliamsNB18}) & Premise: I burst through a set of cabin doors, and fell to the ground. \newline Hypothesis: I burst through the doors and fell down. & Entailment/Neutral/ Contradiction & 353,431 / 39,271 / 9,815 \\
    \midrule
    SICK-E \cite{DBLP:conf/lrec/MarelliMBBBZ14} & Sentence A: A group of kids is playing in a yard and an old man is standing in the background \newline Sentence B: A group of boys in a yard is playing and a man is standing in the background & Entailment/Neutral / Contradiction & 4,439 / 495 / 4,906 \\
    \midrule 
    SNLI \cite{DBLP:conf/emnlp/BowmanAPM15}& Text: A soccer game with multiple males playing. \newline Hypothesis: Some men are playing a sport. &  Entailment/Neutral/ Contradiction & 549,367 / 9,842 / 9,824 \\
    \midrule 
    SST-5  \cite{DBLP:conf/emnlp/SocherPWCMNP13} &  a metaphor for a modern-day urban china searching for its identity .  & very positive/positive/ neutral / negative / very negative & 8,544 / 1,101 / 2,210 \\
    \midrule
    Twitter Financial News Sentiment \footnote{https://huggingface.co/datasets/zeroshot/twitter-financial-news-sentiment}  & "\$BYND - JPMorgan reels in expectations on Beyond Meat https://t.co/bd0xbFGjkT"	& Bearish/ Bullish / Neutral & 8,587 / 956 / 2,388 \\
    \midrule
    Yelp review (Yelp; \citealt{DBLP:conf/nips/ZhangZL15})  & Tonya is super sweet and the front desk people are very helpful	& 1, 2, 3, 4, 5 & 585,000 / 65,000 / 50,000 \\
    \bottomrule
\end{tabularx}
\caption{\mbox{The examples for the given datasets and the original label and statistics before reformatting.}}

\label{tab:appendix:dataset}
\end{table}

\subsection{Dataset Normalization} \label{appendix:data_norm}
We follow the setup of the previous work \cite{DBLP:journals/corr/abs-1904-04096} to convert 5 degrees of sentiment to 3 classes:
for Amazon, APP, and Yelp, we map 1 and 2 to \ti{Negative}, 3 to \ti{Neutral}, and 4 and 5 to \ti{Positive};
for SST-5, we map \ti{very positive} and \ti{positive} to \ti{Positive}, and \ti{negative} and \ti{very negative} to \ti{Negative}.
for TFNS, the original class labels \ti{Bearish} and \ti{Bullish} are mapped to \ti{Negative} and \ti{Positive}.

\clearpage
\subsection{Experimental Setup} \label{appendix:hardness:experiment}
The seeds of training are \{1, 10, 100\}, and the learning rate is 1e-5.
The detailed configuration of \tf{Roberta-Large}, \tf{OPT-350M} and \tf{Flan-T5-Large}, including training are shown at Table \ref{tab:training_config_roberta}, \ref{tab:training_config_opt}, and \ref{tab:training_config_flant5}.
All the experiments are conducted on a single NVIDIA A100.

\FloatBarrier
\begin{table}[htbp]
\small
    \centering
    \begin{tabular}{lccc}
    \toprule
         Datasets & Batch size & Epochs & Seq. length \\
         \midrule
         Amazon       & 16 & 10 & 512 \\
         APP          & 16 & 10 & 256 \\
         MNLI         & 64 & 10 & 128 \\
         SICK-E       & 16 & 10 & 256 \\
         SST-5        & 16 & 10 & 128 \\
         SNLI         & 20 & 5 & 128 \\
         TFNS         & 16 & 10 & 128 \\
         Yelp         & 24 & 5 & 256 \\
    \bottomrule
    \end{tabular}
    \caption{Training configuration of \tf{Roberta-Large}.}
    \label{tab:training_config_roberta}
\end{table}

\begin{table}[htbp]
\small
    \centering
    \begin{tabular}{lccc}
    \toprule
         Datasets & Batch size & Epochs & Seq. length \\
         \midrule
         Amazon       & 6 & 3 & 256 \\
         APP          & 16 & 10 & 256 \\
         MNLI         & 64 & 5 & 128 \\
         SICK-E       & 16 & 10 & 256 \\
         SNLI         & 64 & 5 & 128 \\
         SST-5         & 64 & 10 & 128 \\
         TFNS         & 64 & 10 & 128 \\
         Yelp         &16 & 5 & 256 \\
    \bottomrule
    \end{tabular}
    \caption{Training configuration of \tf{OPT-350M}}
    \label{tab:training_config_opt}
\end{table}

\begin{table}[htbp]
\small
    \centering
    \begin{tabular}{lccc}
    \toprule
         Datasets & Batch size & Epochs & Seq. length \\
         \midrule
         Amazon       & 6 & 3 & 256 \\
         APP          & 6 & 5 & 256 \\
         MNLI         & 12 & 5 & 128 \\
         SICK-E       & 6 & 10 & 256 \\
         SNLI         & 12 & 5 & 128 \\
         SST-5        & 12 & 10 & 128 \\
         TFNS         & 12 & 10 & 128 \\
         Yelp         & 6 & 5 & 256 \\
    \bottomrule
    \end{tabular}
    \caption{Training configuration of \tf{Flan-T5}.}
    \label{tab:training_config_flant5}
\end{table}

\subsubsection{\ti{Neutral}'s hardness in human disagreement} \label{appendix:hardness:human}
\begin{figure}[h]
\centering
\includegraphics[width=0.5\linewidth]{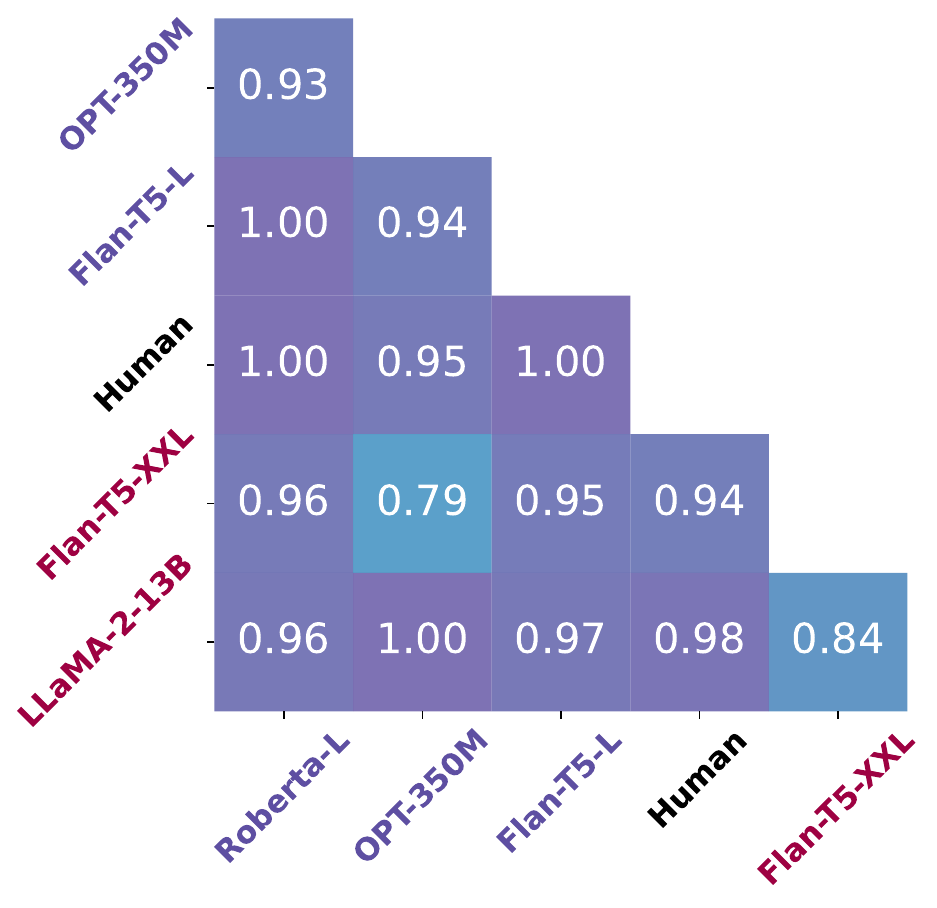}
\caption{Correlation matrix among class-wise F1 scores of three {\color{redpolar}finetuned} models together with two {\color{bluepolar}ICL}s and class-wise human disagreement on MNLI, where the high consistency is noted.}
\label{fig:mnli_corr}
\end{figure}

We present the human variation on the classification as a hardness measurement from human beings.
Table \ref{tab:human_disagreement} presents the distribution of human disagreement of MNLI \cite{DBLP:conf/naacl/WilliamsNB18} and SNLI \cite{DBLP:conf/emnlp/BowmanAPM15}.
The high entropy of \ti{Neutral} reveals its class-wise hardness for humans.
For the convenient comparison with other metrics, we take the negative of the entropy to obtain the positive correlation between the knowledge, as shown in Figure \ref{fig:snli_corr} and \ref{fig:mnli_corr}.

\begin{table}[t]
{\small
    \begin{center}
    \begin{tabular}{lccc}
    \toprule
        \tiny{\backslashbox{Dataset}{Class}} & Entailment & Neutral & Contradiction \\
         \midrule
         MNLI & 0.3202  & \tf{0.4717} & 0.2664 \\
         SNLI & 0.3515  & \tf{0.5175} & 0.2781 \\
         \bottomrule
    \end{tabular}
    \end{center}
}   
    \vspace{-0.8em}
    \caption{Average entropy of annotation distribution for the instances belonging to the same class in MNLI and SNLI. \tf{Bold} indicates the highest entropy score.}
    \label{tab:human_disagreement}
\end{table}

\FloatBarrier

\subsubsection{\ti{Neutral}'s hardness in LLMs} \label{appendix:hardness:llms}
Regarding the hardness of \ti{Neutral} w.r.t. LLMs, we conduct two families of LLMs, \tf{Flan-T5-XXL} \cite{DBLP:journals/jmlr/RaffelSRLNMZLL20} and \tf{LLaMA-2-13B} \cite{DBLP:journals/corr/abs-2307-09288}.
The prompting templates for MNLI and SNLI present as follows:

\begin{tcolorbox}
The prompt for \tf{Flan-T5}: \\
\{\ti{premise}\}. Does this imply \{\ti{hypothesis}\}? options: \\
entailment \\ 
contradiction \\ 
neutral
\end{tcolorbox}

\begin{tcolorbox}
The prompt for \tf{LLaMA-2-13B}: \\
Input: \{\ti{premise}\} Question: Does this imply that \{\ti{hypothesis}\}? Please respond with 'Entailment', 'Contradiction', or 'Neutral'. Result:
\end{tcolorbox}

\begin{table*}[ht]
\centering
\small
\begin{tabular}{l|ccc|ccc}
\toprule
\multirow{2}{*}{\backslashbox{Models}{Datasets}}  & \multicolumn{3}{c|}{\tf{MNLI}} & \multicolumn{3}{c}{\tf{SNLI}} \\
& \tf{Entailment} & \tf{Neutral} & \tf{Contradiction} & \tf{Entailment} & \tf{Neutral} & \tf{Contradiction} \\
\midrule
\tf{Flan-T5-XXL} & 0.90 & 0.87 & 0.94 & 0.90 & 0.88 & 0.93 \\
\tf{LLaMA-2-13B} & 0.51 & 0.28 & 0.50 & 0.41 & 0.39 & 0.55 \\

\bottomrule
\end{tabular}
\caption{F1 scores of in-context learning using  \tf{Flan-T5-XXL} and \tf{LLaMA-2-13B} on MNLI and SNLI. \tf{Bold} indicates the poorest performance across the class. The results are averaged on the seeds \{100, 200, 300\}}
\label{tab:appendix:llm}
\vspace{-0.05in}
\end{table*}

\FloatBarrier

\section{Hardness measurement}
\subsection{Taxonomy of hardness measurement} \label{appendix:measurement:taxonomy}

\begin{figure}[!htbp]
		\centering
		\includegraphics[width=0.7\columnwidth]{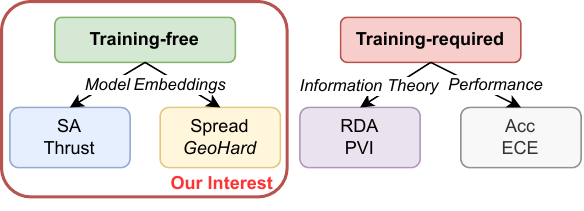}
		\caption{Taxonomy of hardness measurement and the scope of this work.}
		\label{fig:tax}
		\vspace{-1\baselineskip}
\end{figure}

\subsection{SA} \label{appendix:measurement:sa}
The technical steps of SA are as follows:
\begin{enumerate}[label=\arabic*)]
\item Train a Roberta-Large model $\theta$ with $\mathcal{D}_{train}$ and evaluate the model with $D_{test} = \{(X^{test}_i, y^{test}_i)\}$;
\item By randomly masking several consecutive words on $X_i$ and reconstructing $K$ samples with LMs, generate perturbed test dataset $D'_{test} = \{(X'_{ij}, y_i)\}$, where $j = 1, ..., K$;
\item Calculate the confidence for each disturbed input $X'_{ij}$ on the golden label $y_i$ with $\theta$;
\item The sensitivity for each input $X_{i}$ is then defined as the maximum value of the deviation between the confidence values for the original $X_i$ and the corresponding perturbed samples $X'_{ij}$.
\end{enumerate}

For each dataset, sensitivity values are averaged on the three Roberta-Large models trained in Table \ref{tab:hardness_model}.

\subsection{Thrust} \label{appendix:measurement:thrust}
Thrust \cite{DBLP:journals/corr/abs-2307-10442} measures how likely the query to LLMs can be solved by the internal knowledge of the target model, in other words, how necessary the knowledge is needed to propel the model's inference.
There are two essential assumptions of Thrust:
(1) LLMs are expected to well study the given tasks.
(2) Meanwhile, the particular samples deviate from the output embeddings of LLMs, mainly due to insufficient knowledge of LLMs.

We denote the representation function, namely the decoder of UnifiedQA-Flan-T5-Large, as $f_{thrust}(\cdot)$.
We sample a certain number of instances from the datasets, i.e., $D^{sample}$.
Concretely, for the task of sentiment classification, $D^{sample} = 200$, and for the task of natural language inference, $D^{sample} = 600$.
Based on the representation obtained, the samples belonging to the identical classes are grouped together as $\mathcal{G}_l = \{ (f_{thrust}(x_i), y_i,) |  y_i = l \}$, where $(x_i, y_i)$ are the sampled instances, and $l$ is the class index.
Then, each $G_l$ is clustered to $K$ clusters by the k-means algorithm, and each cluster and its corresponding centroid are denoted as $C_{kl}$ and $m_{kl}$, respectively.
Regarding the selection of cluster numbers $K$, we refer to the original setup \cite{DBLP:journals/corr/abs-2307-10442}, i.e., $K = max(ceil(\sqrt[4]{|D^{sample}|}), 3)$.
The seeds for sample selection and clustering initialization are both $\{2, 4, 42, 102, 144\}$, and hence the results are averaged on 25 initial setups.

\begin{equation*}
    s_{thrust}(q) = \norm{ \frac{1}{ N \cdot K } \sum_{l=1}^{N} \sum_{k=1}^{K} \frac{|\mathcal{C}_{kl}|}{\| d_{kl}(q) \|^2 } \cdot \frac{d_{kl} (q)}{\| d_{kl}(q) \|} }
\end{equation*}

where $q$ denotes the query, namely, the test instance, and $d_{kl} = m_{kl} - f(q)$ is a vector pointing from $f(q)$ towards the centroid $m_{kl}$.

The prompts for NLI and SC tasks used on the model are shown as follows:

\begin{tcolorbox}
The prompt for \tf{NLI} tasks: \\
\{\ti{premise}\}. And \{\ti{hypothesis}\}. What is the relationship between these two sentences? Option: Entailment or Neutral or Contradiction. Answer: 
\end{tcolorbox}

\begin{tcolorbox}
The prompt for \tf{SC} tasks: \\
\{sentence\}. Is it a happy review? Answer:
\end{tcolorbox}

\subsection{Spread} \label{appendix:measurement:spread}
Spread aims to measure the instance-level hardness in the few-shot scenario \cite{DBLP:conf/emnlp/ZhaoMM22}.
The idea of Spread is to examine the similarity between training and test instances.
Concretely, if one test sample is close to train samples semantically, it is taken as an easy instance.
We denote the semantic encoder for Spread as $f_{Spread}$.
The distance of one test instance to the training set is defined as the average distance between the test instance to the k-closest training instances.
Let $D^{tr} = \{(x^{tr}_i, y^{tr}_i)\}$ and $D^{te} = \{ (x^{te}_i, y^{te}_i) \}$ denote the training and test sets, respectively.
$x^{tr}_{ik}$ denotes the $k$-th closest training instances to the test instance $x^{te}_i$.
$K_{shot}$ is the number of shots to the training sets, and $d(\cdot, \cdot)$ is the measurement between two data points.

\begin{equation*}
    s_{Spread}(x^{te}_{i}) = \frac{1}{K_{shot}} \sum_{k=1}^{K_{shot}} d( x^{te}_i, x^{tr}_{ik} )
\end{equation*}

\subsection{\ts{PVI}} \label{appendix:measurement:pvi}
Algorithm \ref{appendix:measurement:pvi_algo} presents the procedure of \ts{PVI} \cite{DBLP:conf/icml/EthayarajhCS22}. 
$g'$ is fine-tuned on the original training dataset $\mathcal{D}$, i.e., $\{ (X_i, y_i) | \forall (X_i, y_i) \in \mathcal{D}\}$.
Meanwhile, $g$ is fine-tuned on the null-target pairs $\{ (\varnothing, y_i) | \forall (X_i, y_i) \in \mathcal{D} \}$, where $\varnothing$ is an empty string.

\begin{algorithm}
\caption{\ts{PVI} calculation}\label{appendix:measurement:pvi_algo}
\begin{algorithmic}[1]
\Require a dataset $\mathcal{D} = \{ (X_{1:N}, y_{1:N}) \}$, a model $\mathcal{G}$, and the test instance of ($X^{test}$, $y^{test}$)
\State $g'$ $\leftarrow$ fine-tune $\mathcal{G}$ on $\mathcal{D}$
\State $g$ $\leftarrow$ fine-tune $\mathcal{G}$ on $\{ (\varnothing, y_i) | \forall (X_i, y_i) \in \mathcal{D} \}$
\State \ts{PVI}($X^{test}$, $y^{test}$) $\leftarrow$ $-\log_2 g[\varnothing](y^{test}) + \log_2g'[X^{test}](y^{test})$
\end{algorithmic}
\end{algorithm}

\section{Experimental results} \label{appendix:measurement:result}
As a supplement of Table \ref{tab:corr_table}, the following Table \ref{tab:appendix:main_results} presents the fine-grained numerical values of the golden hardness and different metrics.

\begin{table*}[h!]
\begin{center}
\centering
\resizebox{1.0\textwidth}{!}{%
\begin{tabular}{lcccccccc}
\toprule
\tf{Datasets} & \tf{Class} & \tf{F1}($\downarrow$) & \tf{Sensitivity}($\uparrow$) & \tf{Thrust}($\uparrow$) & \tf{Spread}($\uparrow$)  & \tf{Intra-class} ($\uparrow$) & \tf{Inter-class} ($\uparrow$) & \ti{GeoHard} \\
\midrule
\multirow{3}{*}{\tf{Amazon}}    & Positive &87.6\tiny{$\pm$0.41}&0.1708\tiny{$\pm$0.0121}&0.455\tiny{$\pm$0.004}&0.839&2.837\tiny{$\pm$0.003}&-11.152\tiny{$\pm$0.096}&-8.316\tiny{$\pm$0.096}\\
{}                              & Neutral   &  71.0\tiny{$\pm$0.96}&0.2511\tiny{$\pm$0.0222}&0.575\tiny{$\pm$0.008}&0.842&2.968\tiny{$\pm$0.007}&-6.786\tiny{$\pm$0.056}&-3.818\tiny{$\pm$0.061}\\
{}                              & Negative    & 80.5\tiny{$\pm$0.69}&0.3153\tiny{$\pm$0.0158}&0.513\tiny{$\pm$0.027}&0.844&2.712\tiny{$\pm$0.007}&-9.205\tiny{$\pm$0.071}&-6.493\tiny{$\pm$0.076}\\

\midrule
\multirow{3}{*}{\tf{APP}}      & Positive     & 74.2\tiny{$\pm$0.16}&0.245\tiny{$\pm$0.033}&0.54\tiny{$\pm$0.024}&0.876&5.359\tiny{$\pm$0.012}&-7.28\tiny{$\pm$0.72}&-1.921\tiny{$\pm$0.709}\\
{}                             & Neutral      & 60.1\tiny{$\pm$0.14}&0.1368\tiny{$\pm$0.0062}&0.447\tiny{$\pm$0.014}&0.864&6.251\tiny{$\pm$0.012}&-4.945\tiny{$\pm$0.342}&1.306\tiny{$\pm$0.353}\\
{}                             & Negative     & 73.4\tiny{$\pm$0.8}&0.2724\tiny{$\pm$0.0252}&0.513\tiny{$\pm$0.027}&0.862&5.668\tiny{$\pm$0.05}&-7.24\tiny{$\pm$0.72}&-1.571\tiny{$\pm$0.77}\\
\midrule
\multirow{3}{*}{\tf{MNLI}}    & Entailment    & 91.1\tiny{$\pm$0.12}&0.8233\tiny{$\pm$0.01}&1.503\tiny{$\pm$0.021}&0.837&4.013\tiny{$\pm$0.008}&-0.049\tiny{$\pm$0.005}&3.964\tiny{$\pm$0.012}\\
{}                            & Neutral       & 87.2\tiny{$\pm$0.12}&0.496\tiny{$\pm$0.0085}&1.503\tiny{$\pm$0.021}&0.832&4.037\tiny{$\pm$0.009}&-0.062\tiny{$\pm$0.007}&3.975\tiny{$\pm$0.015}\\
{}                            & Contradiction & 92.8\tiny{$\pm$0.05}&0.6828\tiny{$\pm$0.0142}&1.503\tiny{$\pm$0.021}&0.833&3.989\tiny{$\pm$0.01}&-0.07\tiny{$\pm$0.002}&3.92\tiny{$\pm$0.012}\\
\midrule
\multirow{3}{*}{\tf{SICK-E}}     & Entailment    & 92.9\tiny{$\pm$0.22}&0.8968\tiny{$\pm$0.0231}&1.589\tiny{$\pm$0.07}&0.859&3.112\tiny{$\pm$0.026}&-2.205\tiny{$\pm$0.172}&0.906\tiny{$\pm$0.187}\\
{}                               & Neutral       & 86.8\tiny{$\pm$2.11}&0.3036\tiny{$\pm$0.0431}&1.589\tiny{$\pm$0.07}&0.854&3.471\tiny{$\pm$0.013}&-2.363\tiny{$\pm$0.163}&1.108\tiny{$\pm$0.171}\\
{}                               & Contradiction & 92.4\tiny{$\pm$0.46}&0.9049\tiny{$\pm$0.0106}&1.589\tiny{$\pm$0.07}&0.863&2.197\tiny{$\pm$0.002}&-4.135\tiny{$\pm$0.312}&-1.937\tiny{$\pm$0.312}\\
\midrule
\multirow{3}{*}{\tf{SNLI}}       & Entailment    & 92.6\tiny{$\pm$0.08}&0.8712\tiny{$\pm$0.0155}&1.582\tiny{$\pm$0.011}&0.877&5.64\tiny{$\pm$0.018}&-0.069\tiny{$\pm$0.013}&5.571\tiny{$\pm$0.03}\\
{}                               & Neutral       & 89.2\tiny{$\pm$0.17}&0.6938\tiny{$\pm$0.0205}&1.582\tiny{$\pm$0.011}&0.87&5.645\tiny{$\pm$0.02}&-0.064\tiny{$\pm$0.015}&5.581\tiny{$\pm$0.035}\\
{}                               & Contradiction & 95.3\tiny{$\pm$0.08}&0.5447\tiny{$\pm$0.0188}&1.88\tiny{$\pm$0.179}&0.864&5.596\tiny{$\pm$0.021}&-0.104\tiny{$\pm$0.035}&5.491\tiny{$\pm$0.056}\\
\midrule
\multirow{3}{*}{\tf{SST-5}}      & Positive    & 83.1\tiny{$\pm$0.73}&0.2242\tiny{$\pm$0.0389}&0.821\tiny{$\pm$0.014}&0.828&1.648\tiny{$\pm$0.022}&-7.665\tiny{$\pm$0.366}&-6.017\tiny{$\pm$0.346}\\
{}                               & Neutral     & 53.1\tiny{$\pm$1.55}&0.2347\tiny{$\pm$0.0667}&0.801\tiny{$\pm$0.013}&0.824&1.904\tiny{$\pm$0.01}&-5.014\tiny{$\pm$0.261}&-3.11\tiny{$\pm$0.27}\\
{}                               & Negative    & 75.8\tiny{$\pm$1.7}&0.364\tiny{$\pm$0.0451}&0.764\tiny{$\pm$0.022}&0.83&1.68\tiny{$\pm$0.023}&-7.376\tiny{$\pm$0.416}&-5.696\tiny{$\pm$0.418}\\
\midrule
\multirow{3}{*}{\tf{TFNS}}    & Positive    & 93.0\tiny{$\pm$0.08}&0.3627\tiny{$\pm$0.0585}&0.553\tiny{$\pm$0.035}&0.818&2.25\tiny{$\pm$0.025}&-7.836\tiny{$\pm$0.176}&-5.587\tiny{$\pm$0.191}\\
{}                               & Neutral     & 86.1\tiny{$\pm$0.29}&0.1292\tiny{$\pm$0.0133}&0.523\tiny{$\pm$0.03}&0.806&2.391\tiny{$\pm$0.02}&-6.204\tiny{$\pm$0.301}&-3.813\tiny{$\pm$0.287}\\
{}                               & Negative    & 92.2\tiny{$\pm$0.54}&0.4689\tiny{$\pm$0.0169}&0.752\tiny{$\pm$0.033}&0.819&2.769\tiny{$\pm$0.083}&-7.481\tiny{$\pm$0.163}&-4.712\tiny{$\pm$0.144}\\
\midrule
\multirow{3}{*}{\tf{Yelp}}       & Positive    & 87.9\tiny{$\pm$0.14}&0.0451\tiny{$\pm$0.0021}&0.455\tiny{$\pm$0.006}&0.822&4.043\tiny{$\pm$0.011}&-9.793\tiny{$\pm$0.012}&-5.75\tiny{$\pm$0.016}\\
{}                               & Neutral     & 75.4\tiny{$\pm$0.25}&0.0832\tiny{$\pm$0.0033}&0.395\tiny{$\pm$0.006}&0.819&4.396\tiny{$\pm$0.003}&-6.328\tiny{$\pm$0.009}&-1.931\tiny{$\pm$0.006}\\
{}                               & Negative    & 86.6\tiny{$\pm$0.12}&0.0639\tiny{$\pm$0.0036}&0.455\tiny{$\pm$0.006}&0.811&4.108\tiny{$\pm$0.015}&-9.19\tiny{$\pm$0.018}&-5.082\tiny{$\pm$0.033}\\
\bottomrule
\end{tabular}}
\end{center}

\caption{The class hardness measurement on the tasks containing the undetermined class \ti{Neutral} using SA, Spread, and Thrust to class hardness measurement. $\downarrow$ indicates that the lower value reflects more hardness, while $\uparrow$ indicates that the higher value reflects more hardness.}
\label{tab:appendix:main_results}
\end{table*}

\subsection{\ti{GeoHard}}
\subsubsection{NLI's fine-grained results with different connecting words or phrases} \label{appendix:our_eval:connecting}
Table \ref{tab:dist_complexity_nli} and \ref{tab:biased_gravity_nli} present the measurement of \ti{Distributional complexity} and \ti{Biased gravity}.
Different from the SC datasets, a pair of sentences is in the NLI task.
Therefore, a conjunction word is needed to convert a pair of sentences to a natural sentence.
As shown in Table \ref{tab:dist_complexity_nli} and \ref{tab:biased_gravity_nli}, six conjunctive words or phrases are selected, including \ti{And}, \ti{It is true that}, etc.
As mentioned above, we average the metrics on different conjunctions to measure the NLI sentence pair.

\begin{table*}[hbt!]
\begin{center}
\centering
\resizebox{1.0\textwidth}{!}{%
\begin{tabular}{lccccccccc}
\toprule
\tf{Datasets} & \tf{Class} & \ti{Maybe} & \ti{And} & \ti{Therefore} & \ti{But} & \ti{On the other hand} & \ti{It is true that} & \tf{Average} \\
\midrule
\multirow{3}{*}{\tf{MNLI}}    & Positive    & 4.078\tiny{$\pm$0.003}&4.04\tiny{$\pm$0.008}&3.956\tiny{$\pm$0.024}&4.017\tiny{$\pm$0.029}&4.016\tiny{$\pm$0.001}&3.972\tiny{$\pm$0.05}& 4.013 \\
{}                              & Neutral     &4.099\tiny{$\pm$0.011}&4.061\tiny{$\pm$0.008}&3.976\tiny{$\pm$0.026}&4.037\tiny{$\pm$0.035}&4.056\tiny{$\pm$0.001}&3.993\tiny{$\pm$0.052}& 4.037 \\
{}                              & Negative  & 4.063\tiny{$\pm$0.001}&4.017\tiny{$\pm$0.009}&3.906\tiny{$\pm$0.018}&3.99\tiny{$\pm$0.03}&4.013\tiny{$\pm$0.001}&3.946\tiny{$\pm$0.061}& 3.989 \\
\midrule
\multirow{3}{*}{\tf{SNLI}}      & Positive     &6.256\tiny{$\pm$0.058}&5.738\tiny{$\pm$0.064}&5.314\tiny{$\pm$0.034}&5.44\tiny{$\pm$0.069}&5.513\tiny{$\pm$0.028}&5.579\tiny{$\pm$0.062}& 5.640 \\
{}                             & Neutral      & 6.261\tiny{$\pm$0.057}&5.75\tiny{$\pm$0.061}&5.31\tiny{$\pm$0.043}&5.445\tiny{$\pm$0.067}&5.523\tiny{$\pm$0.029}&5.581\tiny{$\pm$0.065}& 5.645 \\
{}                             & Negative     &6.26\tiny{$\pm$0.059}&5.722\tiny{$\pm$0.06}&5.188\tiny{$\pm$0.019}&5.382\tiny{$\pm$0.072}&5.467\tiny{$\pm$0.046}&5.556\tiny{$\pm$0.055}& 5.596 \\

\midrule
\multirow{3}{*}{\tf{SICK-E}}    & Entailment    &3.521\tiny{$\pm$0.003}&3.066\tiny{$\pm$0.3}&2.537\tiny{$\pm$0.086}&2.705\tiny{$\pm$0.029}&3.445\tiny{$\pm$0.035}&3.397\tiny{$\pm$0.289}& 3.112 \\

{}                            & Neutral       &3.749\tiny{$\pm$0.048}&3.398\tiny{$\pm$0.184}&3.192\tiny{$\pm$0.038}&3.287\tiny{$\pm$0.013}&3.639\tiny{$\pm$0.024}&3.561\tiny{$\pm$0.209}& 3.471 \\
{}                            & Contradiction &2.181\tiny{$\pm$0.027}&2.212\tiny{$\pm$0.061}&2.155\tiny{$\pm$0.029}&2.252\tiny{$\pm$0.017}&2.212\tiny{$\pm$0.016}&2.172\tiny{$\pm$0.038}& 2.197 \\
\bottomrule
\end{tabular}}
\end{center}

\caption{\ti{Intra-clas} metrics of premise and hypothesis concatenated with different conjunctions on three NLI datasets. The results is averaged on three seeds.}
\label{tab:dist_complexity_nli}
\end{table*}

\begin{table*}[hbt!]
\begin{center}
\centering
\resizebox{1.0\textwidth}{!}{%
\begin{tabular}{lccccccccc}
\toprule
\tf{Datasets} & \tf{Class} & \ti{Maybe} & \ti{And} & \ti{Therefore} & \ti{But} & \ti{On the other hand} & \ti{It is true that} & \tf{Average} \\
\midrule
\multirow{3}{*}{\tf{MNLI}}    & Positive    &-0.033\tiny{$\pm$0.019}&-0.039\tiny{$\pm$0.003}&-0.103\tiny{$\pm$0.004}&-0.042\tiny{$\pm$0.019}&-0.027\tiny{$\pm$0.002}&-0.05\tiny{$\pm$0.01}& -0.049
 \\
{}                              & Neutral     &-0.05\tiny{$\pm$0.027}&-0.052\tiny{$\pm$0.003}&-0.101\tiny{$\pm$0.007}&-0.067\tiny{$\pm$0.031}&-0.038\tiny{$\pm$0.004}&-0.065\tiny{$\pm$0.027}&  -0.062\\
{}                              & Negative  &-0.036\tiny{$\pm$0.01}&-0.063\tiny{$\pm$0.006}&-0.168\tiny{$\pm$0.005}&-0.048\tiny{$\pm$0.013}&-0.044\tiny{$\pm$0.002}&-0.058\tiny{$\pm$0.003}& -0.07 \\
\midrule
\multirow{3}{*}{\tf{SNLI}}      & Positive     & -0.011\tiny{$\pm$0.004}&-0.035\tiny{$\pm$0.005}&-0.111\tiny{$\pm$0.018}&-0.068\tiny{$\pm$0.016}&-0.123\tiny{$\pm$0.09}&-0.064\tiny{$\pm$0.009}& -0.069 \\
{}                             & Neutral      & -0.012\tiny{$\pm$0.004}&-0.034\tiny{$\pm$0.008}&-0.092\tiny{$\pm$0.015}&-0.064\tiny{$\pm$0.008}&-0.122\tiny{$\pm$0.084}&-0.058\tiny{$\pm$0.008}&-0.064  \\
{}                             & Negative     &-0.013\tiny{$\pm$0.005}&-0.048\tiny{$\pm$0.013}&-0.164\tiny{$\pm$0.034}&-0.11\tiny{$\pm$0.028}&-0.203\tiny{$\pm$0.188}&-0.087\tiny{$\pm$0.029}&  -0.104\\
\midrule
\multirow{3}{*}{\tf{SICK-E}}    & Entailment    &-1.372\tiny{$\pm$0.199}&-2.822\tiny{$\pm$0.9}&-2.992\tiny{$\pm$0.316}&-2.877\tiny{$\pm$0.317}&-1.335\tiny{$\pm$0.161}&-1.834\tiny{$\pm$0.363}&-2.205  \\
{}                            & Neutral       & -1.614\tiny{$\pm$0.219}&-2.811\tiny{$\pm$0.897}&-3.274\tiny{$\pm$0.368}&-3.027\tiny{$\pm$0.277}&-1.364\tiny{$\pm$0.152}&-2.089\tiny{$\pm$0.385}& -2.363 \\
{}                            & Contradiction & -2.501\tiny{$\pm$0.382}&-5.259\tiny{$\pm$1.564}&-5.631\tiny{$\pm$0.615}&-5.397\tiny{$\pm$0.498}&-2.609\tiny{$\pm$0.308}&-3.411\tiny{$\pm$0.711}& -4.135 \\
\bottomrule
\end{tabular}}
\end{center}

\caption{\ti{Inter-class} metrics of premise and hypothesis concatenated with different conjunctions on three NLI datasets. The results is averaged on three seeds.}
\label{tab:biased_gravity_nli}
\end{table*}

\FloatBarrier

\section{Generalization and Application of \ti{GeoHard}}
\subsection{Theoretical generalization} \label{appendix:general:proof}
Taking the data distribution as one Gaussian distribution $D \sim \mathcal{N}(\mu, \sigma^2)$, the mean of $n$ instances sampled from $D$ follows $\mathcal{N}(\mu, {\sigma^2}/{n})$.
Therefore, the means of the training data and the test data, $\hat{\mu}_{tr}$ and $\hat{\mu}_{te}$, follow $\mathcal{N}(\mu, \sigma^2 / n_{tr})$ and $\mathcal{N}(\mu, \sigma^2 / n_{te})$, where $n_{tr}$ and $n_{te}$ are the size of training and test sets, respectively.
According to Chebyshev's inequality \cite{mitrinovic2013classical}, the following inequalities stand with arbitrary $k \in R_{+}$:

\begin{align*}
  P(|\hat{\mu}_{tr} - \mu| \geq \frac{k \sigma}{\sqrt{n_{tr}}} ) \leq \frac{1}{k^2} \\
  P(|\hat{\mu}_{te} - \mu| \geq \frac{k \sigma}{\sqrt{n_{te}}} ) \leq \frac{1}{k^2} 
\end{align*}

Assuming $n_{tr} \geq n_{te}$ without loss of generality, we combine the two inequalities above and derive:
\begin{align*}
& \frac{2}{k^2} \geq  P(|\hat{\mu}_{tr} - \mu| \geq \frac{k \sigma}{ \sqrt{ n_{tr} } } ) +  P(|\hat{\mu}_{te} - \mu| \geq \frac{k \sigma}{ \sqrt{ n_{te} } } ) \\
& \stackrel{}{\geq} P(|\hat{\mu}_{tr} - \mu| \geq \frac{k \sigma}{ \sqrt{ n_{te} } } ) +  P(|\hat{\mu}_{te} - \mu| \geq \frac{k \sigma}{ \sqrt{ n_{te} } } ) \\
& \geq P(|\hat{\mu}_{tr} - \mu| + |\hat{\mu}_{te} - \mu|\geq \frac{2k \sigma}{ \sqrt{ n_{te} } }) \\
&\geq P(|\hat{\mu}_{tr} - \hat{\mu}_{te}| \geq \frac{2k \sigma}{ \sqrt{ n_{te} } })
\end{align*}

\subsection{\ti{Neutral}'s overfitting} \label{appendix:general:figure}

\begin{figure}[h]
\centering
    \begin{subfigure}[b]{0.5\textwidth}
        \center
        \includegraphics[width=\linewidth]{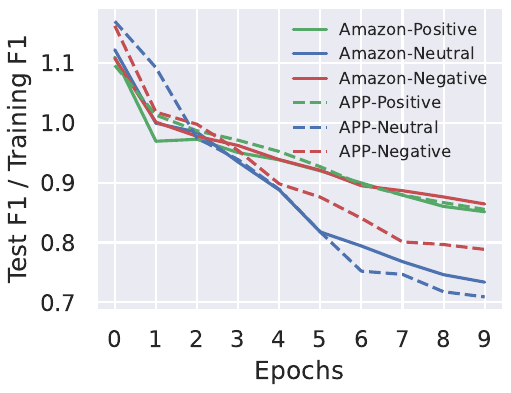}
    \end{subfigure}%
    ~ 
    \begin{subfigure}[b]{0.5\textwidth}
        \includegraphics[width=\linewidth]{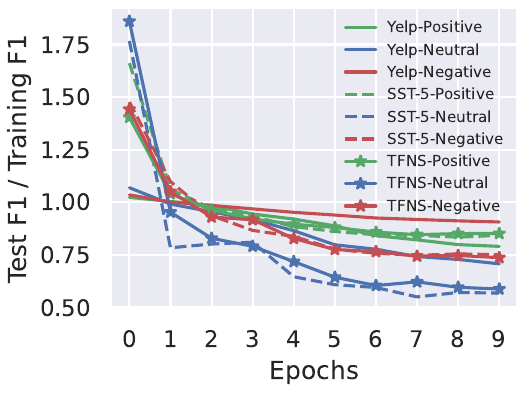}
    \end{subfigure}
    \caption{The ratio between F1 scores on the test and training sets for each training epoch on SC tasks (Left: Amazon and APP; Right: Yelp, SST-5 and TFNS).}
\label{fig:sc_dist_diff}
\end{figure}

\subsection{Empirical Validation on \ti{GeoHard}'s Generalization}
\label{appendix:general:empirical_proof}
As mentioned in the main part of the paper, we also include four datasets, i.e., AG News, Yahoo, Emo, and CARAR, from the tasks of emotion detection and topic classification.
Similarly, we formulate the datasets to balance the number of instances inside each class to achieve class-wise balance.
Specifically, we re-sample 10,000 instances from each class in Yahoo to handle the trade-off between computational efficiency and representativeness.

Trained on Roberta-Large with three seeds $\{1, 10, 100\}$, the performance of four datasets present the class-wise F1 scores in Table \ref{tab:appendix:agnews} - \ref{tab:appendix:yahoo}.

\begin{table}[htbp]
\centering
\begin{tabular}{lcccc}
\toprule
\multicolumn{1}{c}{AG News}  & \ti{World} & \ti{Sports} & \ti{Business} & \ti{Sci/Tech} \\ \midrule
F1 score (\%) & 96.4\tiny{$\pm$0.08} & 99.1\tiny{$\pm$0.07} & 92.7\tiny{$\pm$0.16} & 93.2\tiny{$\pm$0.14} \\
\ti{Intra-class} & 4.464\tiny{$\pm$.016}&3.835\tiny{$\pm$.034}&3.775\tiny{$\pm$.042}&3.978\tiny{$\pm$.070}\\
\ti{Inter-class} &-15.49\tiny{$\pm$.133}&-17.62\tiny{$\pm$.100}&-13.11\tiny{$\pm$.221}&-13.35\tiny{$\pm$.075}\\
\ti{GeoHard}&-11.02\tiny{$\pm$.119}&-13.79\tiny{$\pm$.084}&-9.340\tiny{$\pm$.253}&-9.377\tiny{$\pm$.018}\\
\bottomrule
\end{tabular}
\caption{AG News's class-wise F1 scores trained with Roberta-Large and class-wise hardness measured by \ti{GeoHard}.}
\label{tab:appendix:agnews}
\end{table}

\begin{table}[htbp]
\centering
\begin{tabular}{lccccc}
\toprule
\multicolumn{1}{c}{Emo}  & \ti{Others} & \ti{Happy} & \ti{Sad} & \ti{Angry} \\ \midrule
F1 score (\%) & 82.4\tiny{$\pm$1.09} & 89.8\tiny{$\pm$0.44} & 90.1\tiny{$\pm$1.37} & 91.5\tiny{$\pm$0.94} \\
\ti{Intra-class} & 2.331\tiny{$\pm$.025}&2.286\tiny{$\pm$.024}&2.120\tiny{$\pm$.015}&2.117\tiny{$\pm$.006}\\
\ti{Inter-class} &  -6.841\tiny{$\pm$.081}&-9.781\tiny{$\pm$.147}&-9.521\tiny{$\pm$.186}&-8.278\tiny{$\pm$.428}\\
\ti{GeoHard} & -4.509\tiny{$\pm$.063}&-7.495\tiny{$\pm$.170}&-7.400\tiny{$\pm$.177}&-6.160\tiny{$\pm$.425}\\
\bottomrule
\end{tabular}
\caption{Emo's class-wise F1 scores trained on Roberta-Large and class-wise hardness measured by \ti{Geohard}.}
\label{tab:appendix:emo}
\end{table}

\begin{table}[htbp]
\centering
\begin{tabular}{lcccccc}
\toprule
\multicolumn{1}{c}{CARAR}  & \ti{Sadness} & \ti{Joy} & \ti{Love} & \ti{Anger} & \ti{Fear} & \ti{Superise} \\ \midrule
F1 score (\%) & 90.9\tiny{$\pm$0.49} & 89.1\tiny{$\pm$0.61} & 90.4\tiny{$\pm$0.68} & 93.2\tiny{$\pm$1.25} & 90.1\tiny{$\pm$0.09} & 95.1\tiny{$\pm$0.83} \\
\ti{Intra-class} &1.817\tiny{$\pm$.129}&2.285\tiny{$\pm$.068}&1.656\tiny{$\pm$.013}&1.641\tiny{$\pm$.075}&1.505\tiny{$\pm$.040}&1.675\tiny{$\pm$.007}\\
\ti{Inter-class} &-6.538\tiny{$\pm$.144}&-6.232\tiny{$\pm$.216}&-5.806\tiny{$\pm$.079}&-7.159\tiny{$\pm$.122}&-6.043\tiny{$\pm$.123}&-6.822\tiny{$\pm$.116}\\
\ti{GeoHard} &-4.720\tiny{$\pm$.263}&-3.947\tiny{$\pm$.199}&-4.150\tiny{$\pm$.066}&-5.517\tiny{$\pm$.196}&-4.537\tiny{$\pm$.107}&-5.147\tiny{$\pm$.111}\\
\bottomrule
\end{tabular}
\caption{CARAR's class-wise F1 scores trained on Roberta-Large and class-wise hardness measured by \ti{GeoHard}.}
\label{tab:appendix:carar}
\end{table}

\begin{sidewaystable}
\centering
\scriptsize
\begin{tabular}{lcccccccccc}
\toprule
\multicolumn{1}{c}{Yahoo}  & \ti{0} & \ti{1} & \ti{2} & \ti{3} & \ti{4} & \ti{5} & \ti{6} & \ti{7} & \ti{8} & \ti{9} \\ \midrule
F1 score (\%) & 64.8\tiny{$\pm$0.72} & 77.7\tiny{$\pm$0.28} & 82.0\tiny{$\pm$0.19} & 59.7\tiny{$\pm$0.50} & 87.8\tiny{$\pm$0.64} & 91.9\tiny{$\pm$0.12} & 59.6\tiny{$\pm$0.17} & 76.8\tiny{$\pm$0.30} & 78.8\tiny{$\pm$0.04} & 81.5\tiny{$\pm$0.23} \\
\ti{Intra-class} & 3.08\tiny{$\pm$.014}&2.920\tiny{$\pm$.009}&2.726\tiny{$\pm$.013}&3.416\tiny{$\pm$.014}&2.550\tiny{$\pm$.021}&3.156\tiny{$\pm$.004}&4.257\tiny{$\pm$.042}&4.481\tiny{$\pm$.066}&3.195\tiny{$\pm$.020}&2.814\tiny{$\pm$.027}\\
\ti{Inter-class} &-7.836\tiny{$\pm$.033}&-8.894\tiny{$\pm$.012}&-11.63\tiny{$\pm$.092}&-6.872\tiny{$\pm$.006}&-12.38\tiny{$\pm$.137}&-13.97\tiny{$\pm$.053}&-7.303\tiny{$\pm$.053}&-8.428\tiny{$\pm$.042}&-10.11\tiny{$\pm$.062}&-7.294\tiny{$\pm$.066}\\
\ti{GeoHard} & -4.747\tiny{$\pm$.019}&-5.974\tiny{$\pm$.003}&-8.910\tiny{$\pm$.080}&-3.455\tiny{$\pm$.015}&-9.838\tiny{$\pm$.117}&-10.81\tiny{$\pm$.056}&-3.045\tiny{$\pm$.012}&-3.946\tiny{$\pm$.025}&-6.922\tiny{$\pm$.043}&-4.480\tiny{$\pm$.042}\\
\bottomrule
\end{tabular}
\caption{Yahoo's class-wise F1 scores trained with Roberta-Large and class-wise hardness measured by \ti{GeoHard}. In Yahoo, the class index from 0-9 denotes the classes (\ti{Society\&Culture}, \ti{Science\&Mathematics}, \ti{Health}, \ti{Education\&Reference}, \ti{Computers\&Internet}, \ti{Sports}, \ti{Business\&Finance}, \ti{Entertainment\&Music}, \ti{Family\&Relationships}, \ti{ Politics\&Government})}
\label{tab:appendix:yahoo}
\end{sidewaystable}

\clearpage

\subsection{Other metrics beyond training-free methods}
\label{appendix:other_metrics}

Here, we include two training-based methods to further validate the existence of class-wise hardness.
One is \ts{PVI} \cite{DBLP:conf/icml/EthayarajhCS22} and the other is data cartography \cite{swayamdipta-etal-2020-dataset}.
\ts{PVI} has been introduced in Appendix \ref{appendix:measurement:pvi}.
Data cartography focuses on the behavior of the model on data instances during training, referred to as training dynamics.
This includes two metrics for each instance: the model's \ti{confidence} in the correct class and the \ti{variability} of this confidence across epochs.
Data points characterized by high confidence and low variability are considered easy.
In Table \ref{tab:training_based_corr_table}, we observe that \ts{PVI} can well model the hardness of the classes. 
Moreover, we also notice the correlation between class-wise hardness and the metrics from training dynamics.
These results further validates the existence of class-wise hardness through a training-based way.

\begin{table*}[t]
\centering
\small
\begin{tabular}{l|ccccc|ccc|c}
\toprule
{\multirow{2}{*}{\backslashbox{\makebox[1em][l]{Metric}}{\makebox[2.4em][l]{Dataset}}}} & \multicolumn{5}{c|}{\tf{SC}} & \multicolumn{3}{c|}{\tf{NLI}} & \multirow{2}{*}{\shortstack{Macro Avg.}} \\
&\tf{Amazon} & \tf{APP} & \tf{SST-5} & \textbf{TFNS} & \textbf{Yelp} & \textbf{MNLI}  &\textbf{SNLI} & \textbf{SICK-E} &  \\
\midrule
\ts{PVI} & 0.985 & 0.9825 & 0.9808 & 1.0000 & 1.0000 & 0.9805 & 0.9628 & 0.9973 & 0.9861 \\
Confidence & 0.6966 & 0.9947 & 0.9959 & 0.997 & 0.7183 & 0.9721 & 0.9897 & 0.7478 & 0.8890 \\
Variablity & -0.8755 & -0.5244 & -0.1256 & 0.8683 & -0.7595 & -0.9991 & -0.9949 & -0.0117 & -0.4278 \\
\bottomrule
\end{tabular}
\caption{Pearson's correlation coefficients between class-wise hardness measurement and class-wise F1 scores.
The metrics include \ts{PVI} and two metrics from training dynamics, i.e., confidence and variability.
 }
\label{tab:training_based_corr_table}
\end{table*}

\subsection{\ti{GeoHard}'s Application}
\label{appendix:general:application_demostrations}
The following demonstrates two examples of the demonstration applied in the ICL on Dynasent \cite{DBLP:conf/acl/PottsWGK20}. Precisely, the upper and the lower demonstrations are \textsc{2P+1Neu+1N} and \textsc{1P+2Neu+1N}, respectively.

\begin{tcolorbox}
Sentence: This place is fine.i love this place, the staff is great the food is great and the atmosphere is great. \\
Sentiment: pos1 \\
\#\#\#\#\# \\
Sentence: The casino has some of the lowest house-edge blackjack you will find anywhere. \\
Sentiment: positive \\ 
\#\#\#\#\# \\
Sentence: Too bad they only had one available spot that day, it was an appointment at 4:30pm, fortunately for me that is the least busiest time so I was in and out. \\
Sentiment: neutral \\
\#\#\#\#\# \\
Sentence: I went to the ticket counter. I wasn't going to the ticket counter after the show demanding a refund, but I certainly wouldn't go again.  \\
Sentiment: negative \\
\#\#\#\#\# \\
Sentence: \{input\} \\
Sentiment: 

\end{tcolorbox}

\begin{tcolorbox}
Sentence: I tried a new place. I definitely recommend this place if you are looking for some good chinese food, and I definitely will be coming back. \\
Sentiment: positive \\
\#\#\#\#\# \\
Sentence: It was cool. It is set up like a lounge, but it has a dinky dancefloor, and music that is WAY TOO LOUD for a place that has a lounge setup. \\
Sentiment: mixed \\
\#\#\#\#\# \\
Sentence: So I'll give this one just one store. \\
Sentiment: neutral \\
\#\#\#\#\# \\
Sentence: I went to the ticket counter. I wasn't going to the ticket counter after the show demanding a refund, but I certainly wouldn't go again.  \\
Sentiment: negative \\
\#\#\#\#\# \\
Sentence: \{input\} \\
Sentiment: \\
\end{tcolorbox}

\clearpage
\section{Artifacts and Packages}
\label{appendix:artifacts}
The details of the datasets, major packages, and existing models are listed in Table \ref{tab:tools}.

\definecolor{mygray}{gray}{.9}
\begin{table*} 
    \centering
    \resizebox{\textwidth}{!}{
    \begin{tabular} 
    {llll} 
       \toprule %
        \textbf{Artifacts/Packages} & \textbf{Citation} & \textbf{Link} & \textbf{License}\\ 
        \rowcolor{mygray} \multicolumn{4}{c}{\textit{Artifacts(datasets/benchmarks). }}\\
        Amazon & \cite{DBLP:conf/emnlp/KeungLSS20} & \url{https://huggingface.co/datasets/amazon_reviews_multi} & \href{https://docs.opendata.aws/amazon-reviews-ml/license.txt}{LICENSE} \\
        APP & \cite{DBLP:conf/sigsoft/GranoSMVCP17} & \url{https://huggingface.co/datasets/app_reviews} & Missing\\
        MNLI & \cite{DBLP:conf/naacl/WilliamsNB18} & \url{https://huggingface.co/datasets/multi_nli} & MIT License\\
        SICK-E & \cite{DBLP:conf/lrec/MarelliMBBBZ14} & \url{https://huggingface.co/datasets/sick} & CC-by-NC-SA-3.0\\
        SNLI & \cite{DBLP:conf/emnlp/BowmanAPM15} & \url{https://huggingface.co/datasets/snli} & CC-by-4.0\\
        SST-5 & \cite{DBLP:conf/emnlp/SocherPWCMNP13} & \url{https://huggingface.co/datasets/SetFit/sst5} & Missing\\
        TFNS & N/A & \url{https://huggingface.co/datasets/zeroshot/twitter-financial-news-sentiment} & MIT License\\
        Yelp & \cite{DBLP:conf/nips/ZhangZL15} & \url{https://huggingface.co/datasets/yelp_review_full} & \href{https://s3-media3.fl.yelpcdn.com/assets/srv0/engineering_pages/bea5c1e92bf3/assets/vendor/yelp-dataset-agreement.pdf}{LICENSE}\\
         Dynasent & \cite{DBLP:conf/acl/PottsWGK20} & \url{https://github.com/cgpotts/dynasent} & Apache License 2.0\\
        \rowcolor{mygray} \multicolumn{4}{c}{\textit{Packages}}\\
        PyTorch & \cite{paszke-etal-2019-pytorch} & \url{https://pytorch.org/} & BSD-3 License\\
        transformers & \cite{wolf2019huggingface} & \url{https://huggingface.co/transformers/v2.11.0/index.html} & Apache License 2.0\\
        numpy & \cite{harris-etal-2020-array} & \url{https://numpy.org/} & BSD License \\
        pandas & \cite{mckinney-2010-data} & \url{https://pandas.pydata.org/} & BSD 3-Clause License \\
        matplotlib & \cite{hunter2007matplotlib} & \url{https://matplotlib.org/} & BSD compatible License\\
        umap & \cite{McInnes2018} & \url{https://github.com/lmcinnes/umap} & BSD 3-Clause License \\
       \rowcolor{mygray} \multicolumn{4}{c}{\textit{Models}}\\
        E5-Large-v2 & \cite{DBLP:journals/corr/abs-2212-03533} & \url{https://huggingface.co/intfloat/e5-large-v2} & MIT License\\
        GTE-Large & \cite{DBLP:journals/corr/abs-2308-03281} & \url{https://huggingface.co/thenlper/gte-large} & MIT License\\
        bge-large-en-v1.5 & \cite{DBLP:journals/corr/abs-2309-07597} & \url{https://huggingface.co/BAAI/bge-large-en-v1.5} & MIT License\\
        RoBERTa & \cite{DBLP:journals/corr/abs-1907-11692} & \url{https://huggingface.co/docs/transformers/model_doc/roberta} & MIT License\\
        Flan-T5 & \cite{DBLP:journals/jmlr/RaffelSRLNMZLL20} & \url{https://huggingface.co/docs/transformers/model_doc/flan-t5} & Apache-2.9 \\
        OPT & \cite{DBLP:journals/corr/abs-2205-01068} & \url{https://huggingface.co/facebook/opt-2.7b} & \href{https://github.com/facebookresearch/metaseq/blob/main/projects/OPT/MODEL_LICENSE.md}{LICENSE}\\
        LLaMA-v2 & \cite{DBLP:journals/corr/abs-2307-09288} & \url{https://huggingface.co/docs/transformers/model_doc/llama2} & \href{https://ai.meta.com/llama/license/}{LICENSE}\\
      \bottomrule %
    \end{tabular}
    }
    \caption{Details of datasets, major packages, and existing models we use. The datasets we reconstructed or revised and the code/software we provide are under the MIT License. }
    \label{tab:tools}
\end{table*}

\end{document}